\documentclass[sn-basic,12pt]{sn-jnl_MODIFIED}

\usepackage{graphicx}%
\usepackage{multirow}%
\usepackage{amsmath,amssymb,amsfonts}%
\usepackage{amsthm}%
\usepackage{mathrsfs}%
\usepackage[title]{appendix}%
\usepackage{xcolor}%
\usepackage{textcomp}%
\usepackage{manyfoot}%
\usepackage{booktabs}%
\usepackage{algorithm}%
\usepackage{algorithmicx}%
\usepackage{algpseudocode}%
\usepackage{listings}%
\usepackage{mathtools}
\usepackage{palatino}
\usepackage{siunitx}
\usepackage{bbm}
\usepackage{tcolorbox}

\newcommand{\argmaxF}{\mathop{\mathrm{argmax}}\limits}

\algnewcommand\algorithmicinput{\textbf{Initialize:}}
\algnewcommand\INPUT{\item[\algorithmicinput]}

\theoremstyle{thmstyleone}%
\theoremstyle{thmstyletwo}%

\theoremstyle{thmstylethree}%

\begin{document}

\title[Article Title]{\textbf{Hierarchical mixtures of Unigram models for short text clustering: The role of Beta-Liouville priors}}


\author*[1]{\fnm{Massimo} \sur{Bilancia}}\email{massimo.bilancia@uniba.it}
\author[2]{\fnm{Samuele} \sur{Magro}}\email{magrosamuele1999@gmail.com}

\affil[1]{\orgdiv{Department of Precision and Regenerative Medicine and Jonian Area (DiMePRe-J)}, \orgname{University of Bari Aldo Moro}, \orgaddress{\street{Piazza Giulio Cesare 11}, \city{Bari}, \postcode{70124}, \country{Italy}}}
\affil[2]{\orgdiv{Sirio Astronomical Observatory}, \orgname{Grotte di Castellana srl}, \orgaddress{\street{Piazzale Anelli}, \city{Castellana Grotte}, \postcode{70013}, \country{Italy}}}


\abstract{
This paper presents a variant of the Multinomial mixture model tailored to the unsupervised classification of short text data. While the Multinomial probability vector is traditionally assigned a Dirichlet prior distribution, this work explores an alternative formulation based on the Beta-Liouville distribution, which offers a more flexible correlation structure than the Dirichlet. We examine the theoretical properties of the Beta-Liouville distribution, with particular focus on its conjugacy with the Multinomial likelihood. This property enables the derivation of update equations for a \textsf{CAVI} (Coordinate Ascent Variational Inference) algorithm, facilitating approximate posterior inference of the model parameters. In addition, we introduce a stochastic variant of the \textsf{CAVI} algorithm to enhance scalability. The paper concludes with empirical examples demonstrating effective strategies for selecting the Beta-Liouville hyperparameters.
}




\maketitle

\thispagestyle{empty}
\maketitle
\newpage
\section{Introduction}\label{sec1}

In recent years, natural language processing (NLP) techniques--defined as the application of computational and statistical methods to analyze and synthesize natural language and speech--have experienced remarkable growth. Notably, models based on the transformer architecture, such as BERT \citep{devlin_bert_2019,yvon_transformers_2023}, have emerged as state-of-the-art solutions in NLP, driving the adoption of neural language models in numerous NLP applications \citep{wu_survey_2024}. Despite this paradigm shift, traditional statistical language models, which define a probability distribution over the vocabulary of a corpus, retain significant appeal and utility. These models often provide statistical advantages over contemporary alternatives, particularly for smaller datasets, and are generally more computationally efficient. Additionally, they offer superior interpretability, as they produce probability distributions over term occurrences that can be effectively summarized and visualized \citep{rosenfeld_two_2000,jurafsky_speech_2024}.

Among statistical language models, the $n$-gram model is the most general, estimating the probability of a word based on the preceding $n-1$ words. Typically, these probabilities are estimated via maximum likelihood estimation, with adjustments to address sparsity and undersampling \citep{zitouni_backoff_2007}. When $n = 1$, the model simplifies to a Unigram model, where the probability of each word in a sequence is treated as independent of the others. This simplification allows the probability of a sequence to be expressed as the product of the marginal probabilities of its words. Such a representation corresponds to a ``bag-of-words'' (BOW) approach, which treats text as an unordered collection of words. Moreover, this approach establishes a direct correspondence between word probabilities and the probabilities of a Multinomial distribution, wherein only word frequencies in the sequence are considered.

In the NLP literature, a Multinomial probability distribution over vocabulary terms is commonly referred to as a topic. Extending this concept, we assume the existence of $G$ distinct topics within a corpus, with each document's thematic content associated with one specific topic. Since document-topic assignments are unknown, unsupervised classification can be formulated by assuming a generative model--a mixture of Unigrams--where $G$ Multinomial distributions represent the different topics \citep{nigam_text_2000}. In this context, Bayesian estimation is often employed, with Dirichlet distributions acting as priors for the Multinomial probabilities. This hierarchical model structure is both intuitive and advantageous, as it addresses a notable limitation of the Multinomial distribution: low counts can result in unreliable probability estimates, and the phenomenon of word burstiness is insufficiently modeled. The Dirichlet prior alleviates these issues by introducing extra-Multinomial variation in the marginal distribution, thereby accounting for deviations from repeated Bernoulli trials \citep{bouguila_count_2011,bakhtiari_latent_2016,anderlucci_mixtures_2020}.

While the Dirichlet-Multinomial composition is computationally convenient, the Dirichlet distribution is often criticized for its rigid covariance structure, which enforces negative off-diagonal elements \citep{bouguila_clustering_2008}. However, correlation is essential for accurately modeling text structure and can enhance the performance of text categorization. To address these limitations, the Beta-Liouville distribution has emerged as an alternative prior in statistical language modeling \citep{koochemeshkian_advancing_2023}. Similar to the Dirichlet, the Beta-Liouville distribution serves as a conjugate prior for the Multinomial distribution but introduces an additional parameter that enables a more flexible covariance structure. This flexibility mitigates the negative correlation constraints imposed by the Dirichlet prior and, in certain cases, allows for positive correlations. Although positive correlations are less relevant to our study, as demonstrated in Section \ref{sec:dataexamples}, the enhanced flexibility of the Beta-Liouville prior remains advantageous.

In this paper, we introduce a fully Bayesian approach to the Beta-Liouville Multinomial mixture model. This novel formulation adopts a hierarchical structure, with posterior inference carried out through a variational method in which the posterior distribution is approximated by a surrogate composed of independent components. For the proposed model, we derive the update equations for the coordinate ascent variational inference (\textsf{CAVI}) algorithm \citep{blei_variational_2017}, which maximizes a lower bound on the marginal likelihood with respect to a set of variational parameters. We further show how a stochastic variant of the \textsf{CAVI} algorithm can be implemented \citep{hoffman_stochastic_2013}, wherein the update equations do not depend of the full dataset. This significantly improves scalability by substituting key quantities with noisy yet unbiased and computationally efficient Monte Carlo estimates based on individual data points. Since the Dirichlet distribution is a special case of the Beta-Liouville distribution, the primary objective of this work is to compare the Dirichlet-Multinomial and Beta-Liouville Multinomial models in settings characterized by data sparsity, such as text corpora consisting of short and semantically ambiguous documents (e.g., abstracts). Our goal is to assess whether the additional parameterization afforded by the Beta-Liouville distribution yields substantial benefits in unsupervised classification tasks.

The structure of this paper is as follows. Section \ref{sec:generative} presents the generative processes underlying both the Dirichlet-Multinomial and Beta-Liouville Multinomial mixture models, including the derivation of the Beta-Liouville distribution and its role as a generalization of the Dirichlet. This section also examines the key properties that support Bayesian inference with this prior, such as its inclusion in the exponential family and its conjugacy with the Multinomial likelihood. Section \ref{sec:variationallearning} outlines the variational inference framework, providing detailed derivations of the update equations required to compute variational posterior estimates of the model parameters, and introduces the stochastic variational inference algorithm. Section \ref{sec:dataexamples} evaluates the performance of Beta-Liouville Multinomial mixtures in comparison to Dirichlet-Multinomial mixtures, using the latter as a baseline. The evaluation is carried out on two short-text corpora, with unsupervised classification accuracy serving as the primary performance metric. Topic coherence is also assessed when the model is employed for dimensionality reduction. Finally, Section \ref{sec:discussion_conclusions} discusses the results and outlines potential directions for future research.

\section{The generative model}\label{sec:generative}

Denote by $\mathbb V$ a vocabulary of terms over a corpus of documents, where $p = \vert\mathbb V \vert$ represents the number of distinct terms. In a formal context, the standard assumption is that the data-generating process can be modeled as a generative probabilistic framework that produces infinitely exchangeable streams of terms \citep{Gelman2013}. This implies that two finite sequences of identical length, differing only in the order of terms, are generated with the same probability and correspond to the same bag-of-words (BOW) representation. Specifically, a document is represented as a vector of counts:
\begin{equation}
	\pmb y = (y_1, \ldots, y_p)^\top,
\end{equation}
where $y_\ell$, $\ell = 1,\ldots,p$, denotes the number of occurrences of the $\ell$-th term in the vocabulary $\mathbbm V$. Infinite exchangeability ensures that the probability of observing any finite stream of terms, regardless of length, can be expressed as the product (invariant to permutations) of the corresponding marginal probabilities. These conditions naturally lead to the Unigram language model, wherein the likelihood of the vector of counts $\pmb y$ assumes the well-known Multinomial form:
\begin{equation}\label{eq:multinomial}
	p(\pmb y \vert \pmb\pi) = \frac{\left(\sum_{\ell = 1}^p  y_\ell\right)!}{\prod_{\ell}^p y_\ell!}\prod_{\ell = 1}^p \pi_\ell^{y_\ell},
\end{equation}
where $\pmb\pi = (\pi_1, \ldots, \pi_p)^\top \in \mathbb R^p$ is the vector of Multinomial parameters, which must satisfy the constraints $\pi_\ell > 0$ for $\ell = 1,\ldots,p$ and $\sum_{\ell = 1}^p \pi_\ell = 1$. The normalizing constant in \eqref{eq:multinomial} is irrelevant for inference, as sequences differing solely in the order of terms yield identical estimates.

The additional assumption is that each document in the corpus belongs to one and only one of $G$ distinct thematic categories ($g = 1,\ldots,G$), referred to as topics. These are represented by an unobserved latent variable $\pmb z = (z_1,\ldots,z_G)^\top$ indicating group membership, where $z_g = 1$ if $\pmb y$ originates from topic $g$, and $z_g = 0$ otherwise. Each thematic category is characterized by a specific Multinomial distribution over the vocabulary of terms, parameterized by vectors $\pmb\pi_g$. The data-generating process is thus defined as:
\begin{equation*}
	p(\pmb y, \pmb z\vert \pmb{\mathbbm P}) = p(\pmb y \vert \pmb{\mathbbm P}, \pmb z) p(\pmb z) = p(\pmb y \vert \pmb\pi_{\pmb z}) \lambda_{\pmb z},
\end{equation*}
where $\lambda_{\pmb z} \equiv p(\pmb z)$, and $\pmb{\mathbbm P}^\top \coloneq (\pmb \pi_1, \ldots, \pmb \pi_G)^\top \in \mathbb R^{G \times p}$ denotes the matrix of topic-specific Multinomial parameters. Since only $\pmb y$ is observed, the marginal distribution of counts is given by \citep{gormley_mixture_2018}:
\begin{equation}\label{eq:mixunigrams}
	p(\pmb y \vert \pmb{\mathbbm P}) = \sum_{\pmb z} \lambda_{\pmb z}  p(\pmb y \vert \pmb\pi_{\pmb z})  = \sum_{g=1}^G \lambda_g p(\pmb y \vert \pmb\pi_g).
\end{equation}

The model in \eqref{eq:mixunigrams} corresponds to a mixture of Multinomials, although it is commonly referred to as a mixture of Unigrams in the text categorization literature. Bayesian inference and the deconvolution of the components in this model require the specification of appropriate priors for the weight vector $\pmb\lambda = (\lambda_1,\ldots,\lambda_G)^\top$ and the probability vectors $\pmb\pi_g$. While the Dirichlet distribution is a natural choice for these priors, the following development introduces the theory necessary to employ the Beta-Liouville distribution as a generalized prior for the Multinomial vectors.

\subsection{Dirichlet-Multinomial mixture}\label{sec:dirmultisec}

In the specific case of Dirichlet-Multinomial mixtures, the generative model \eqref{eq:mixunigrams} for a corpus of $n$ documents can be expressed hierarchically as follows:
\begin{eqnarray}
	\pmb y_i \vert \pmb{\mathbbm P}, z_{ig} = 1 & \stackrel{\textrm{ind.}}{\sim} & \operatorname{Multinomial}_{p}(\pmb \pi_g),\quad i=1,\ldots,n, \label{eq:AV1} \\
	\pmb z_i \vert \pmb \lambda & \stackrel{\textrm{ind.}}{\sim} & \operatorname{Multinoulli}_{G}(\pmb \lambda),\quad i=1,\ldots,n, \label{eq:AV2} \\
	\pmb \pi_g \vert \theta & \stackrel{\textrm{ind.}}{\sim} & \operatorname{Dirichlet}_p(\mathbbm 1_p\theta),\quad g=1,\ldots,G, \label{eq:AV3} \\
	\pmb \lambda \vert \psi & \sim & \operatorname{Dirichlet}_G(\mathbbm 1_G\psi), \label{eq:AV4}
\end{eqnarray}
where $\theta, \psi > 0$, $\mathbbm 1 = (1, 1, \ldots, 1)^\top$ denotes a vector of ones of appropriate dimension, and $\operatorname{Multinoulli}_{G}$ refers to a Multinomial distribution for a single trial. As previously described, the matrix:  
\begin{equation*}
	\pmb{\mathbbm P} = \begin{pmatrix}
		\pmb\pi_1^\top \\
		\vdots \\
		\pmb\pi_G^\top
	\end{pmatrix} \in \mathbb R^{G\times p},
\end{equation*}
where the generic element is denoted as $\pi_{g\ell}$ for $g = 1,2,\ldots,G$ and $\ell = 1,2,\ldots,p$, contains $G$ distinct distributions over the vocabulary of terms $\mathbbm V$ along its rows. Each of these distributions represents a specific topic; however, the actual distribution governing the data generating process for any given document in the corpus remains unknown. Semantically, each document corresponds to one of the $G$ distributions, though the topic label identifying this distribution is unobserved.

In \cite{bilancia_variational_2023}, the standard Coordinate Ascent Variational Inference (\textsf{CAVI}) approach for the posterior inference of the model specified by \eqref{eq:AV1}--\eqref{eq:AV3} is examined. Additionally, \cite{bilancia_stochastic_2024} presents a stochastic variant of \textsf{CAVI} for the same model. In this approach, the key variables estimated at each iteration of the algorithm--ordinarily computed by traversing the entire dataset--are replaced by a noisy but unbiased and computationally efficient estimate based on a single data point \citep{hoffman_stochastic_2013}. This stochastic variational inference method enhances scalability and computational efficiency while introducing an intrinsic random component into the optimization surface, which often helps to avoid suboptimal local minima where deterministic algorithms may become trapped.

\subsection{Multivariate Liouville distribution of second kind}

In this and the following Subsections, we will omit any reference to the latent variable $\pmb z$ and focus on the form of the prior distribution for the Multinomial parameters $\pmb\pi$. A random vector $\pmb X = (X_1, \ldots, X_p)^\top$ is said to follow a Liouville distribution if and only if \citep{gupta_multivariate_1987, gupta_generalized_1996, gupta_multivariate_1999, wesolowski_remarks_1993, gupta_covariance_2001}:  
\begin{equation*}
	\pmb X \stackrel{D}{=} r \pmb U,
\end{equation*}
where $\pmb U \sim \mathrm{Dir}_p(\alpha_1, \ldots, \alpha_p)$ and $r$ is a continuous random variable, independent of $\pmb U$, with distribution function $F$ referred to as the generator. It can be shown that if $r$ admits a density $f$, then the Liouville distribution also has the following density with respect to the Lebesgue measure on $\mathbb R^{p}$:
\begin{equation}\label{eq:liouvillemain}
	p(\pmb x \vert \alpha_1, \ldots, \alpha_p) = \prod_{\ell = 1}^p \frac{x_\ell^{\alpha_\ell - 1}}{\Gamma(\alpha_\ell)} \frac{\Gamma(\alpha_0)}{\left(\sum_{\ell = 1}^p x_\ell\right)^{\alpha_0 - 1}} f\left(\sum_{\ell = 1}^p x_\ell \right),
\end{equation}
where $\alpha_0 = \alpha_1 + \cdots + \alpha_p$, and the support of this distribution is the simplex:
\begin{equation}
	\mathcal D_{p} = \left\{ (x_1, \ldots, x_p) \;\vert\; x_\ell \geq 0, \sum_{\ell = 1}^p x_\ell \leq a \right\},
\end{equation}
if and only if $f$ has support on the interval $(0, a)$. In this case, $\pmb X$ is said to have a Liouville distribution of the second kind. Conversely, the distributions of the first kind arise when the generating variable $r$ has unbounded support, which will not be considered further in the following.

The mixed moments of order $s$ for the distribution defined by the density \eqref{eq:liouvillemain} have the following expression \citep{fang_symmetric_2018}:
\begin{equation*}
	\operatorname{E}\left(\prod_{\ell = 1}^p X_\ell^{s_\ell} \right) =\frac{\mu_s}{{\alpha_0^{[s]}}} \prod_{\ell = 1}^p \alpha_\ell^{[s_\ell]},
\end{equation*}
where:
\begin{itemize}
	\item $s_\ell \geq 0$ and $s = \sum_{\ell = 1}^p s_\ell$.
	\item $\alpha_\ell^{[s_\ell]} =: \alpha_\ell (\alpha_\ell + 1) \cdots (\alpha_\ell + s_\ell - 1)$.
	\item $\mu_s = \operatorname{E}(r^s)$.
\end{itemize}

In particular, simple algebraic manipulations reveal that for $\ell = 1, \ldots, p$:
\begin{equation}\label{eq:mean_liouville}
	\operatorname{E}\left(X_\ell \right) = \mu_1 \frac{\alpha_\ell}{\alpha_0},
\end{equation}
and:
\begin{equation}\label{eq:var_liouville}
	\operatorname{Var}(X_\ell)  =  \mu_2 \frac{\alpha_\ell(\alpha_\ell + 1)}{\alpha_0(\alpha_0 + 1)} - \mu_1^2 \frac{\alpha_\ell^2}{\alpha_0^2}.
\end{equation}

Finally, the correlation structure is described by the following expression ($\ell, \ell' \in \{1, \ldots, p\}$ with $\ell \neq \ell'$):
\begin{equation}\label{eq:cov_liouville}
	\operatorname{Cov}(X_\ell, X_{\ell'}) = \frac{\alpha_\ell \alpha_{\ell'}}{\alpha_0} \left(\frac{\mu_2}{\alpha_0 + 1} - \frac{\mu_1^2}{\alpha_0} \right).
\end{equation}

\subsection{Beta-Liouville distribution}\label{sec:betaliouvilledef}

The density of a multivariate Liouville distribution can be reformulated by introducing the function:  
\begin{equation*}
	g(r) = \frac{\Gamma(\alpha_0)}{r^{\alpha_0 - 1}} f(r),
\end{equation*}  
where $f(r)$ and $g(r)$ share the same support. With this definition, the probability density of a multivariate Liouville distribution can be rewritten as:  
\begin{equation}
	\prod_{\ell = 1}^p \frac{x_\ell^{\alpha_\ell - 1}}{\Gamma(\alpha_\ell)} g\left( \sum_{\ell = 1}^p x_\ell \right).
\end{equation}  

The function $g(r)$ is called a density generator because the density of the generator $f$ can be expressed as follows:  
\begin{equation*}
	f(r) = \frac{r^{\alpha_0 - 1}}{\Gamma(\alpha_0)} g(r).
\end{equation*}  

The case of interest is when $r \sim \textrm{Beta}(\alpha, \beta)$, i.e., the generating random variable has a Beta density, given by:  
\begin{equation*}
	f(r) = \frac{1}{B(\alpha, \beta)} r^{\alpha - 1}(1-r)^{\beta - 1}, \quad 0 \leq r \leq 1.
\end{equation*}  

In this particular case, the density generator has the expression:  
\begin{align*}
	g(r) & = \frac{\Gamma(\alpha_0)}{r^{\alpha_0 - 1}} f(r)  
	= \frac{\Gamma(\alpha_0)}{r^{\alpha_0 - 1}} \frac{1}{B(\alpha, \beta)} r^{\alpha - 1}(1-r)^{\beta - 1}  \\
	& = \frac{\Gamma(\alpha_0)}{B(\alpha, \beta)} r^{\alpha - 1} r^{1 - \alpha_0} (1 - r)^{\beta - 1}  \\
	& = \frac{\Gamma(\alpha_0)}{B(\alpha, \beta)} r^{\alpha - \alpha_0}(1-r)^{\beta - 1}, \quad 0 \leq r \leq 1.
\end{align*}  

With this expression, the multivariate density is given by:  
\begin{equation*}
	p(\pmb x \vert \alpha_1, \ldots, \alpha_{p-1}, \alpha_p, \alpha, \beta) = \prod_{\ell = 1}^p \frac{x_\ell^{\alpha_\ell - 1}}{\Gamma(\alpha_\ell)}\frac{\Gamma(\alpha_0)}{B(\alpha, \beta)} r^{\alpha - \alpha_0}(1-r)^{\beta - 1},
\end{equation*}  
which can be rewritten by taking $r = \sum_{\ell = 1}^p x_\ell$ as the following density on the $(p+1)$-dimensional probability simplex $\mathcal S_{p+1}$ \citep{bouguila_clustering_2008,bouguila_count_2011,bouguila_hybrid_2012,bakhtiari_latent_2016,bourouis_markov_2021}:  
\begin{equation}
	p(\pmb x \vert \alpha_1, \ldots, \alpha_p, \alpha, \beta) = \frac{\Gamma(\alpha_0)}{B(\alpha, \beta)} \prod_{\ell = 1}^p \frac{x_\ell^{\alpha_\ell - 1}}{\Gamma(\alpha_\ell)}\left( \sum_{\ell = 1}^p x_\ell \right)^{\alpha - \alpha_0}\left(1 - \sum_{\ell = 1}^p x_\ell\right)^{\beta - 1}.
\end{equation}

This distribution is called Beta-Liouville and will be denoted as:  
\begin{equation*}
	\pmb X \sim \operatorname{BL}_{p+1}(\alpha_1, \ldots, \alpha_p, \alpha, \beta).
\end{equation*}  

Typically, the notation $\operatorname{BL}_{p}(\alpha_1, \ldots, \alpha_p, \alpha, \beta)$, with index $p$, is used to emphasize that this distribution is described on the $\mathcal D_p$ simplex. However, we have preferred to adopt a symbolism consistent with the Dirichlet distribution, as the Beta-Liouville distribution has realizations on the $\mathcal S_{p+1}$ probability simplex described by the coordinates $
(x_1, \ldots, x_p, 1 - \sum_{\ell = 1}^p x_\ell) \in \mathcal S_{p+1}$.   
It is indeed a generalization of the Dirichlet distribution, obtained as a special case when $\alpha = \alpha_0$.  
In fact, under this assumption, the Beta function can be rewritten as:  
\begin{equation*}
	B(\alpha_0, \beta) = \frac{\Gamma(\alpha_0) \Gamma(\beta)}{\Gamma(\alpha_0 + \beta)},
\end{equation*}  
and the multivariate density becomes:  
\begin{equation*}
	p(x_1, \ldots, x_p, x_{p+1} \vert \alpha_1, \ldots, \alpha_p, \beta) = \frac{\Gamma(\alpha_1 + \cdots + \alpha_p + \beta)}{\Gamma(\beta)} \prod_{\ell = 1}^p \frac{x_\ell^{\alpha_\ell - 1}}{\Gamma(\alpha_\ell)}( \sum_{\ell = 1}^p x_\ell)^{\alpha_0 - \alpha_0}x_{p+1}^{\beta - 1},
\end{equation*}  
with $x_{p+1} = 1 - \sum_{\ell = 1}^p x_\ell$, which simplifies to:  
\begin{equation*}
	p(x_1, \ldots, x_p, x_{p+1} \vert \alpha_1, \ldots, \alpha_p, \beta) = \frac{\Gamma(\alpha_1 + \cdots + \alpha_p + \beta)}{\Gamma(\alpha_1) \cdots \Gamma(\alpha_p) \Gamma(\beta)} \prod_{\ell = 1}^p x_\ell^{\alpha_\ell - 1}x_{p+1}^{\beta - 1}.
\end{equation*}  

This corresponds to the density of a $\operatorname{Dir}_{p+1}(\alpha_1, \ldots, \alpha_p, \beta)$ distribution.  
Thus, to model a probability vector $\pmb\pi = (\pi_1, \ldots, \pi_p)^\top$ on the probability simplex $\mathcal S_p$, we must use a distribution from the family $\operatorname{BL}_{p}(\alpha_1, \ldots, \alpha_{p-1}, \alpha, \beta)$, where the dimension is indexed according to the convention we have chosen.

\subsection{The Beta-Liouville as a conjugate prior of the Multinomial likelihood}\label{sec:conjugate_BL}

Since the Beta-Liouville distribution will be exclusively employed to determine an appropriate prior distribution for the Multinomial parameter vector $\pmb\pi$, we shall consistently use $\pmb\pi = (\pi_1,\ldots,\pi_p)^\top$ as the argument of the distribution in the subsequent discussion. Under these assumptions, the posterior distribution can be expressed as follows:
\begin{align*}
	&\quad\; p(\pmb \pi \vert \alpha_1,\ldots,\alpha_{p-1},\alpha,\beta, \pmb y)  \propto\\ 
	& \propto p(\pmb y \vert \pmb \pi) p(\pmb \pi\vert \alpha_1,\ldots,\alpha_{p-1},\alpha,\beta)  \\
	&\propto \prod_{\ell = 1}^{p-1}\pi_\ell^{y_\ell}\left( 1 - \sum_{\ell = 1}^{p-1} \pi_\ell\right)^{y_p}\times \prod_{\ell = 1}^{p-1}\pi_\ell^{\alpha_\ell - 1}\left(\sum_{\ell = 1}^{p-1}\pi_\ell \right)^{\alpha-\alpha_0}\left(1 - \sum_{\ell = 1}^{p-1}\pi_\ell\right)^{\beta-1}  \\
	& = \prod_{\ell = 1}^{p-1}\pi_\ell^{\alpha_\ell + y_\ell - 1}\left(\sum_{\ell = 1}^{p-1}\pi_\ell \right)^{\alpha-\alpha_0}\left(1 - \sum_{\ell = 1}^{p-1}\pi_\ell\right)^{\beta+ y_p -1}.
\end{align*} 

We thus define $\alpha_\ell' = \alpha_\ell + y_\ell \textrm{ for } \ell = 1,\ldots,p-1$, and $\beta' = \beta + y_p$, from which it follows directly that: 
$ \alpha_\ell = \alpha_\ell' - y_\ell$ and that:
\begin{align*}
	\alpha - \alpha_0 & = \alpha - \left( \alpha_1' - y_1 + \cdots + \alpha_{p-1}' - y_{p-1}\right)  \\
	& = \alpha - \left(\alpha_1'+\cdots+\alpha_{p-1}' \right) + (y_1 + \cdots+y_{p-1})  \\
	& = \alpha  + \sum_{\ell = 1}^{p-1} y_\ell - \sum_{\ell = 1}^{p-1} \alpha_\ell'.
\end{align*}

We can then rewrite the posterior distribution as:
\begin{equation*}
p(\pmb \pi \vert \alpha_1',\ldots,\alpha_{p-1}',\alpha',\beta', \pmb y)  \propto \prod_{\ell = 1}^{p-1}\pi_\ell^{\alpha_\ell' - 1}\left(\sum_{\ell = 1}^{p-1}\pi_\ell \right)^{\alpha'-\alpha_0'}\left(1 - \sum_{\ell = 1}^{p-1}\pi_\ell\right)^{\beta'-1},
\end{equation*}
with:
\begin{align*}
	\alpha_\ell' & = \alpha_\ell + y_\ell,\quad \ell = 1,\ldots,p-1, \\
	\beta' & =  \beta + y_p, \\
	\alpha_0' & = \sum_{\ell = 1}^{p-1} \alpha_\ell', \\
	\alpha' & = \alpha  + \sum_{\ell = 1}^{p-1} y_\ell.
\end{align*}

It is therefore immediate that:
\begin{equation}
\pmb \pi \vert \alpha_1', \ldots, \alpha_{p-1}',\alpha',\beta' ,\pmb y \sim \operatorname{BL}_p(\alpha_1', \ldots, \alpha_{p-1}',\alpha',\beta'),
\end{equation}
demonstrating the conjugacy. It is worth noting that the Beta-Liouville distribution is not the canonical conjugate prior distribution for the Multinomial. Specifically, the Multinomial probability mass function can be expressed in the form of an exponential family in its canonical representation--i.e., parameterized in terms of the natural parameters and presented in its minimal form \citep{brown_fundamentals_1986}:
\begin{equation}\label{eq:exponential}
p(\pmb y \vert \pmb \pi) = h(\pmb y) \exp\left\{\pmb \pi^\top t(\pmb y) - a_{\textrm{lik}}(\pmb \pi)  \right\},
\end{equation}
where $a(\pmb\pi)$ represents the log-partition function. In this case, the canonical conjugate prior distribution can be derived from the distribution \eqref{eq:exponential} and expressed as the following $(k+1)$-dimensional exponential family \citep{blei_build_2014, efron_exponential_2022}: 
\begin{equation}\label{eq:canonical_conj}
p(\pmb\pi \vert \pmb \nu) = h_c (\pmb \pi) \exp \left\{\pmb \nu_1^\top \pmb \pi  + \nu_2 \left(-a_{\textrm{lik}}(\pmb \pi)  \right) - a_{\textrm{conj}}(\pmb \nu) \right\},
\end{equation}
which has as its natural parameter ($\nu_2 = \nu_2^\top$ is a scalar):
\begin{equation*}
\pmb \nu^\top = (\pmb \nu_1, \nu_2).
\end{equation*}

It is well established that the Dirichlet distribution corresponds to the canonical conjugate prior \eqref{eq:canonical_conj} for the Multinomial likelihood. However, it is important to note that the conjugate prior of a given likelihood is not necessarily unique. As demonstrated above, the Beta-Liouville distribution constitutes an alternative conjugate (non-canonical) prior for the Multinomial likelihood.

\subsection{The Beta-Liouville as an exponential family}
Given that the Beta-Liouville distribution generalizes the Dirichlet distribution, it is unsurprising that it also belongs to the exponential family. The relationships derivable from this property are particularly advantageous for posterior inference via a variational approach. To proceed, we first rewrite the Beta-Liouville distribution in its classical exp-log form:
\begin{align*}
	&\quad\; p(\pmb \pi \vert \alpha_1,\ldots,\alpha_{p-1},\alpha,\beta) = \\
	& =  \frac{\Gamma(\alpha_0)\Gamma(\alpha + \beta)}{\Gamma(\alpha)\Gamma(\beta)\prod_{\ell=1}^{p-1}\Gamma(\alpha_\ell)} \prod_{\ell = 1}^{p-1}\pi_\ell^{\alpha_\ell - 1}\left(\sum_{\ell = 1}^{p-1}\pi_\ell \right)^{\alpha-\alpha_0}\left(1 - \sum_{\ell = 1}^{p-1}\pi_\ell\right)^{\beta-1} \\
	& = \exp\left\{\log\Gamma(\alpha_0) + \log\Gamma(\alpha + \beta) - \log\Gamma(\alpha) - \log\Gamma(\beta) - \sum_{\ell = 1}^{p-1}\log\Gamma(\alpha_\ell) \right.  \\
	& \left. + \sum_{\ell = 1}^{p-1}(\alpha_\ell - 1)\log\pi_\ell + (\alpha - \alpha_0)\log\sum_{\ell = 1}^{p-1}\pi_\ell + (\beta - 1)\log\left(1 - \sum_{\ell = 1}^{p-1} \pi_\ell \right)\right\}  \\
	& = \exp\left\{\sum_{\ell = 1}^{p-1}(\alpha_\ell - 1)\log\pi_\ell + (\alpha - \alpha_0)\log\sum_{\ell = 1}^{p-1}\pi_\ell + (\beta - 1)\log\left(1 - \sum_{\ell = 1}^{p-1} \pi_\ell \right) \right. \\
	& - \left.\left(\underbrace{\log\Gamma(\alpha) + \log\Gamma(\beta) + \sum_{\ell = 1}^{p-1}\log\Gamma(\alpha_\ell) - \log\Gamma(\sum_{\ell=1}^{p-1}\alpha_\ell) - \log\Gamma(\alpha + \beta)}_{\textrm{log-partition function $a(\alpha_1,\ldots, \alpha_{p-1}, \alpha, \beta)$}} \right)\right\}.
\end{align*}

Now:
\begin{align*}
	&\quad\;\sum_{\ell = 1}^{p-1}(\alpha_\ell - 1)\log\pi_\ell + (\alpha - \alpha_0)\log\sum_{\ell = 1}^{p-1}\pi_\ell + (\beta - 1)\log\left(1 - \sum_{\ell = 1}^{p-1} \pi_\ell \right)  \nonumber \\
	& = \sum_{\ell = 1}^{p-1}\alpha_\ell \log \pi_\ell - \sum_{\ell = 1}^{p-1}\log\pi_\ell + \alpha \log \sum_{\ell = 1}^{p-1}\pi_\ell - \sum_{\ell = 1}^{p-1}\alpha_\ell \log\sum_{\ell = 1}^{p-1} \pi_\ell \nonumber \\
	& + \beta\log\left(1 - \sum_{\ell = 1}^{p-1} \pi_\ell \right)  - \log\left(1 - \sum_{\ell = 1}^{p-1} \pi_\ell \right)  \nonumber \\
	& = \sum_{\ell = 1}^{p-1}\alpha_\ell \left[\log\pi_\ell - \log\sum_{\ell = 1}^{p-1}\pi_\ell \right] + \alpha \log\sum_{\ell = 1}^{p-1}\pi_\ell + \beta\log\left(1 - \sum_{\ell = 1}^{p-1} \pi_\ell \right)  \nonumber \\
	& \underbrace{- \sum_{\ell = 1}^{p-1}\log\pi_\ell - \log\left(1 - \sum_{\ell = 1}^{p-1} \pi_\ell \right)}_{\textrm{absorbed in $h(\pmb\pi)$}}.
\end{align*}

In the representation as an exponential family, the corresponding minimal sufficient statistics are given by:
\begin{align}
	S1:\; & \left[\log\pi_\ell - \log\sum_{\ell = 1}^{p-1}\pi_\ell \right], \quad \ell = 1,\ldots,p-1 \\
	S2:\; & \log\sum_{\ell = 1}^{p-1}\pi_\ell \\
	S3:\; & \log\left(1 - \sum_{\ell = 1}^{p-1} \pi_\ell \right)
\end{align}
and the corresponding natural parameters are:
\begin{equation}\label{eq:naturalparameters}
	S1:\alpha_\ell, \quad \ell = 1,\ldots,p-1, \qquad S2: \alpha, \qquad S3: \beta
\end{equation}
	
The vector of expected values for the minimal sufficient statistics can be obtained by computing the partial derivative of the log-partition function with respect to the corresponding natural parameter. Consequently:
\begin{align}
	\operatorname{E}_{p_{\pmb \pi}}\left( \log\pi_\ell - \log\sum_{\ell = 1}^{p-1}\pi_\ell \right)  & = \frac{\partial}{\partial \alpha_\ell}a(\alpha_1,\ldots,\alpha_{p-1},\alpha, \beta) = \frac{\Gamma'(\alpha_\ell)}{\Gamma(\alpha_\ell)} - \frac{\Gamma'(\sum_{\ell = 1}^{p-1}\alpha_\ell)}{\Gamma(\sum_{\ell = 1}^{p-1}\alpha_\ell)} \nonumber \\ 
	& = \Psi(\alpha_\ell) - \Psi(\sum_{\ell = 1}^{p-1}\alpha_\ell), \quad \ell =1, \ldots, p-1.
\end{align}

Proceeding analogously for $\alpha$:
\begin{align}
	\operatorname{E}_{p_{\pmb \pi}}\left( \log\sum_{\ell = 1}^{p-1}\pi_\ell \right) & = \frac{\partial}{\partial \alpha}a(\alpha_1,\ldots,\alpha_{p-1},\alpha, \beta) = \frac{\Gamma'(\alpha)}{\Gamma(\alpha)} - \frac{\Gamma'(\alpha + \beta)}{\Gamma(\alpha + \beta)} \nonumber \\
	& = \Psi(\alpha) - \Psi(\alpha + \beta),
\end{align}
and also for $\beta$, we obtain:
\begin{align}
	\operatorname{E}_{p_{\pmb \pi}}\left( \log\left[1 - \sum_{\ell = 1}^{p-1} \pi_\ell \right] \right) & = \frac{\partial}{\partial \beta}a(\alpha_1,\ldots,\alpha_{p-1},\alpha, \beta) = \frac{\Gamma'(\beta)}{\Gamma(\beta)} - \frac{\Gamma'(\alpha + \beta)}{\Gamma(\alpha + \beta)}\nonumber \\ 
	& = \Psi(\beta) - \Psi(\alpha + \beta),\label{eq:blsufficient1}
\end{align}
where $\Psi(z)$ denotes the Digamma function, the logarithmic derivative of the Gamma function. It is also evident from the expressions we have derived that:
\begin{align}
	\operatorname{E}_{p_{\pmb \pi}}\left(\log \pi_\ell \right) & =
	\operatorname{E}_{p_{\pmb \pi}}\left(\log \sum_{\ell = 1}^{p-1} \pi_\ell \right) + \Psi(\alpha_\ell) - \Psi(\sum_{\ell = 1}^{p-1}\alpha_\ell)\nonumber\\
	& = \Psi(\alpha_\ell) - \Psi(\sum_{\ell = 1}^{p-1}\alpha_\ell) + \Psi(\alpha) - \Psi(\alpha + \beta), \quad \ell=1,\ldots,p-1.\label{eq:blsufficient2}
\end{align}

\subsection{Properties of the moments}

In a frequentist approach to inference, the composition of the Multinomial distribution with a Beta-Liouville prior necessitates making the marginal distribution of the counts explicit by integrating out the Multinomial parameter vector $\pmb \pi$. The form of this marginal distribution is well-established in the literature and is given by the following expression:
\begin{equation}\label{eq:multinomial_BL}
	p(\pmb y \vert \alpha_1',\ldots,\alpha_{p-1}',\alpha', \beta') = \frac{\Gamma\left(\sum_{\ell = 1}^p y_\ell + 1\right)}{\prod_{\ell = 1}^p\Gamma(y_\ell + 1)} \times \frac{\Gamma(\alpha_0) \Gamma(\alpha + \beta) \Gamma(\alpha')\Gamma(\beta')\prod_{\ell = 1}^{p-1} \Gamma(\alpha_\ell')}
	{\Gamma(\alpha_0') \Gamma(\alpha' + \beta') \Gamma(\alpha)\Gamma(\beta)\prod_{\ell = 1}^{p-1} \Gamma(\alpha_\ell)},
\end{equation}
where the superscript assigned to the parameters retains the same meaning as in Subsection \ref{sec:conjugate_BL}. The distribution in \eqref{eq:multinomial_BL} is referred to as the Beta-Liouville-Multinomial (abbreviated as BLM$_p$) and should not be confused with the generalized Dirichlet distribution discussed in \cite{connor_concepts_1969}. However, as anticipated, this distribution encompasses the Dirichlet-Multinomial distribution discussed in \cite{Mosimann1962} as a special case. Specifically, when $\alpha = \alpha_0$, the second factor in \eqref{eq:multinomial_BL} simplifies as follows (since $\alpha_0' = \alpha'$):  
\begin{align}\label{eq:factor_MBL}
	&\quad\; \frac{\Gamma(\alpha_0) \Gamma(\alpha + \beta) \Gamma(\alpha')\Gamma(\beta')\prod_{\ell = 1}^{p-1} \Gamma(\alpha_\ell')}
	{\Gamma(\alpha_0') \Gamma(\alpha' + \beta') \Gamma(\alpha)\Gamma(\beta)\prod_{\ell = 1}^{p-1} \Gamma(\alpha_\ell)} 
	 = \frac{\Gamma(\alpha) \Gamma(\alpha + \beta) \Gamma(\alpha')\Gamma(\beta')\prod_{\ell = 1}^{p-1} \Gamma(\alpha_\ell')}
	{\Gamma(\alpha') \Gamma(\alpha' + \beta') \Gamma(\alpha)\Gamma(\beta)\prod_{\ell = 1}^{p-1} \Gamma(\alpha_\ell)} \nonumber\\
	& = \frac{\Gamma(\alpha_0 + \beta) \Gamma(\beta')\prod_{\ell = 1}^{p-1} \Gamma(\alpha_\ell')}{\Gamma(\alpha_0' + \beta')\Gamma(\beta)\prod_{\ell = 1}^{p-1} \Gamma(\alpha_\ell)} 
	= \frac{\Gamma(\alpha_0 + \beta) \Gamma(\beta + y_p )\prod_{\ell = 1}^{p-1} \Gamma(\alpha_\ell + y_\ell)}{\Gamma(\sum_{\ell = 1}^{p-1}\alpha_\ell' +  \beta + y_p)\Gamma(\beta)\prod_{\ell = 1}^{p-1} \Gamma(\alpha_\ell)}  \nonumber\\
	& = \frac{\Gamma(\alpha_0 + \beta) \Gamma(\beta + y_p )\prod_{\ell = 1}^{p-1} \Gamma(\alpha_\ell + y_\ell)}{\Gamma(\alpha_0  + \beta + \sum_{\ell = 1}^p y_\ell)\Gamma(\beta)\prod_{\ell = 1}^{p-1} \Gamma(\alpha_\ell)}.
\end{align}

By setting $\beta \equiv \alpha_p$, we reformulate \eqref{eq:factor_MBL} as:
\begin{equation*}
\frac{\Gamma(\sum_{\ell = 1}^p \alpha_\ell) \prod_{\ell = 1}^{p} \Gamma(\alpha_\ell + y_\ell)}{\Gamma(\sum_{\ell = 1}^p \alpha_\ell + \sum_{\ell = 1}^p y_\ell)\prod_{\ell = 1}^{p} \Gamma(\alpha_\ell)},
\end{equation*}
obtaining the well-known expression for the Dirichlet-Multinomial distribution (DM$_p$), which is precisely the result of composing a Multinomial likelihood with the Dirichlet distribution::
\begin{equation}\label{eq:dirichlet_multinomial}
 p(\pmb y \vert \alpha_1,\ldots,\alpha_{p-1},\alpha_p)  = 
	 \frac{\Gamma\left(\sum_{\ell = 1}^p y_\ell + 1\right)\Gamma(\sum_{\ell = 1}^p \alpha_\ell)}{\Gamma( \sum_{\ell = 1}^p y_\ell + \sum_{\ell = 1}^p \alpha_\ell)}\prod_{\ell = 1}^p\frac{\Gamma(\alpha_\ell + y_\ell)}{\Gamma({\alpha_\ell)}\Gamma(y_\ell + 1)}.
\end{equation}

The moments for the marginal distribution in \eqref{eq:dirichlet_multinomial} are well-known. In particular, the variance of each marginal component exceeds the corresponding variance of the components of a Multinomial distribution, as it is augmented by a factor that depends on an overdispersion parameter. This factor accounts for extra heterogeneity not expected in the pure Multinomial model (see, e.g., \citeauthor{corsini_dealing_2022}, \citeyear{corsini_dealing_2022}).

For the marginal distribution BLM$_p$ in \eqref{eq:multinomial_BL}, the expression for the moments is not readily available. However, since in a Bayesian framework we do not need to integrate over the Multinomial parameters--given that the Multinomial likelihood is combined with the prior distribution in a hierarchical structure--we can instead examine the moments of the BL$_p$ prior, whose expression is well-documented in the literature. Specifically, from the expressions \eqref{eq:mean_liouville}--\eqref{eq:cov_liouville}, we can derive that for $\ell, \ell' \in \{1, \ldots, p-1\}$, with $\ell \neq \ell'$:
\begin{align}
\operatorname{E}(\pi_\ell) & = \frac{\alpha}{\alpha +\beta}\frac{\alpha_\ell}{\alpha_0}, \\
\operatorname{Var}(\pi_\ell) & =  \frac{\alpha(\alpha + 1)}{(\alpha + \beta)(\alpha + \beta + 1)}\left[\frac{\alpha_\ell(\alpha_\ell + 1)}{\alpha_0(\alpha_0 + 1)} \right] - \left(\frac{\alpha}{\alpha + \beta} \right)^2\frac{\alpha_\ell^2}{\alpha_0^2}, \\
\operatorname{Cov}(\pi_\ell, \pi_{\ell'}) & = \frac{\alpha_\ell\alpha_{\ell'}}{\alpha_0(\alpha_0 + 1)}\frac{\alpha(\alpha + 1)}{(\alpha + \beta)(\alpha + \beta + 1)} - \left(\frac{\alpha}{\alpha +\beta}  \right)^2\frac{\alpha_\ell\alpha_{\ell'}}{\alpha_0^2}\label{eq:eqcorrbl}.
\end{align}

The fact that the variance of each marginal component of the BL$_p$ distribution exceeds the corresponding marginal variance of the $\operatorname{Dir}_p$ distribution is inherent to its construction. Additionally, it is well-known that the Dirichlet distribution induces a negative correlation (repulsion) between pairs of components of the random vector. However, as evident from \eqref{eq:eqcorrbl}, this constraint does not apply to the BL$_p$ prior, whose components can exhibit both positive and negative correlations in pairs. Consequently, the Beta-Liouville distribution offers a more flexible correlation structure compared to the Dirichlet distribution, while still being supported on the same probability simplex, at the cost of one additional parameter. It is also worth noting that the literature often incorrectly states that the Beta-Liouville distribution has two more parameters than the Dirichlet distribution.

\subsection{Bayesian hierarchical modelling of Beta-Liouville Multinomial mixtures}

The natural extension of the hierarchical model \eqref{eq:AV1}-\eqref{eq:AV4} to incorporate a Beta-Liouville prior for the Multinomial parameters is:
\begin{eqnarray}
	\pmb y_i \vert \pmb{\mathbbm P}, z_{ig} = 1 & \stackrel{\mathsf{ind.}}{\sim} & \operatorname{Multinomial}_{p}(\pmb \pi_g), \!\!\!\qquad\qquad i=1,2,\ldots,n \label{eq:BAV1}, \\
	\pmb z_i \vert \pmb \lambda & \stackrel{\mathsf{ind.}}{\sim} & \operatorname{Multinoulli}_{G}(\pmb \lambda),  \;\qquad\qquad i=1,2,\ldots,n \label{eq:BAV2},\\
	\pmb \pi_g \vert \alpha_1,\ldots,\alpha_{p-1},\alpha,\beta & \stackrel{\mathsf{ind.}}{\sim} & \operatorname{BL}_p(\alpha_1,\ldots,\alpha_{p-1},\alpha,\beta),\!\quad g=1,2,\ldots,G,  \label{eq:BAV3} \\
	\pmb \lambda\vert\psi  &\sim & \operatorname{Dirichlet}_G(\mathbbm 1_G\psi),\label{eq:BAV4}
\end{eqnarray}
where $\alpha_1, \ldots, \alpha_{p-1}, \alpha, \beta, \psi > 0$. It is also important to note that the joint distribution of the observable counts and latent indicators is given by the following expression:
\begin{equation}\label{eq:indicators_with_hidden}
	p(\pmb y_i,  \pmb z_i  \vert\pmb{\mathbbm P}) = \prod_{g=1}^G \left(\lambda_g\prod_{\ell = 1}^p \pi_{g\ell}^{y_{i\ell}}\right)^{z_{ig}}.
\end{equation}

The unnormalized posterior distribution of the latent parameters can be factorized as:
\begin{eqnarray}\label{eq:posterior}
	p(\pmb{\mathbbm P}, \pmb z_{1:n}, \pmb \lambda \vert \pmb y_{1:n},\alpha_1, \ldots, \alpha_{p-1}, \alpha, \beta, \psi) & \propto &  
	p(\pmb y_{1:n}, \pmb z_{1:n}, \pmb{\mathbbm P}, \pmb \lambda \vert \alpha_1, \ldots, \alpha_{p-1}, \alpha, \beta, \psi)  \nonumber \\
	& = & p(\pmb y_{1:n}\vert \pmb z_{1:n}, \pmb{\mathbbm P}, \pmb \lambda, \alpha_1, \ldots, \alpha_{p-1}, \alpha, \beta, \psi) \nonumber \\
	& \times & p(\pmb{\mathbbm P}, \pmb z_{1:n}, \pmb \lambda \vert \alpha_1, \ldots, \alpha_{p-1}, \alpha, \beta, \psi)  \nonumber\\
	& =  & p(\pmb y_{1:n} \vert \pmb{\mathbbm P}, \pmb z_{1:n})
	p(\pmb z_{1:n} \vert \pmb\lambda) \nonumber \\
	& \times &  p(\pmb{\mathbbm P} \vert \alpha_1, \ldots, \alpha_{p-1}, \alpha, \beta)
	p(\pmb \lambda\vert \psi),
\end{eqnarray}
where $\pmb y_{1:n} = (\pmb y_1, \pmb y_2, \ldots, \pmb y_n)$ denotes a matrix of column vectors, and $\pmb z_{1:n}$ is defined similarly. The posterior distribution cannot be expressed in closed form, as the marginal likelihood of the model is clearly intractable:
\begin{align}\label{eq:marglik}
	& p(\pmb y_{1:n} \vert \alpha_1, \ldots, \alpha_{p-1}, \alpha, \beta, \psi) = \nonumber \\
	& =  \prod_{i=1}^n\int \sum_{\pmb z_i} p(\pmb y_i \vert \pmb{\mathbbm P}, \pmb z_i)
	p(\pmb z_i \vert \pmb\lambda) 
	p(\pmb{\mathbbm P} \vert \alpha_1, \ldots, \alpha_{p-1}, \alpha, \beta)
	p(\pmb \lambda\vert \psi) d\pmb{\mathbbm P} d\pmb \lambda.
\end{align}

\section{Bayesian computations with posterior variational learning} \label{sec:variationallearning}
\subsection{Variational learning}
Variational learning for intractable posteriors approximates the posterior distribution by employing a set of mutually independent variational distributions:  
\begin{equation}
	q_V(\pmb z_{1:n}, \pmb{\mathbbm P}, \pmb \lambda\vert \pmb y_{1:n},  \pmb \gamma_{1:n}, \pmb \phi_{1:G}, \pmb \eta) = q(\pmb \lambda \vert \pmb \eta)\prod_{g=1}^G q(\pmb \pi_g\vert \pmb \phi_g)\prod_{i=1}^n q(\pmb z_i\vert \pmb \gamma_i).
\end{equation}

This formulation relies on a set of variational parameters. By leveraging Jensen's inequality, a lower bound for the marginal likelihood of the model can be derived \citep{jordan_introduction_1999}:
\begin{align}\label{eq:jensen}
	\log p(\pmb y_{1:n} \vert \alpha_1, \ldots, \alpha_{p-1}, \alpha, \beta, \psi) & \geq \operatorname{E}_{q_V}\left[\log  p(\pmb y_{1:n}, \pmb z_{1:n}, \pmb{\mathbbm P}, \pmb \lambda \vert \alpha_1, \ldots, \alpha_{p-1}, \alpha, \beta, \psi)\right] \nonumber \\
	&  - \operatorname{E}_{q_V}\left[\log q_V(\pmb z_{1:n}, \pmb{\mathbbm P}, \pmb \lambda\vert \pmb y_{1:n},  \pmb \gamma_{1:n}, \pmb \phi_{1:G}, \pmb \eta)  \right].
\end{align}

If we abbreviate the second term of the inequality \eqref{eq:jensen} as $\operatorname{ELBO}(q_V)$ (evidence lower bound), the notation emphasizes that the ELBO depends on the choice of the variational distribution $q_V$. Using this abbreviation, we can express the fundamental identity of variational inference \citep{Blei2003,blei_variational_2017,zhang_advances_2019} as:
\begin{align}\label{eq:kldiv}
	& \log p(\pmb y_{1:n} \vert \alpha_1, \ldots, \alpha_{p-1}, \alpha, \beta, \psi)  = \mathsf{ELBO}(q_V) \nonumber \\
	+ &\; \mathsf{ KL }\left( q_V(\pmb z_{1:n}, \pmb{\mathbbm P}, \pmb \lambda\vert \pmb y_{1:n},  \pmb \gamma_{1:n}, \pmb \phi_{1:G}, \pmb \eta) \lvert\rvert p(\pmb{\mathbbm P}, \pmb z_{1:n}, \pmb \lambda \vert \pmb y_{1:n},\alpha_1, \ldots, \alpha_{p-1}, \alpha, \beta, \psi) \right).
\end{align}

Since the reverse Kullback-Leibler (KL) divergence in \eqref{eq:kldiv} is always non-negative and the marginal likelihood is a fixed quantity, minimizing the KL divergence is equivalent to maximizing the $\mathsf{ELBO}$, thereby achieving the tightest possible lower bound for the marginal log-likelihood. In this framework, the estimated variational posterior minimizes its distance to the true posterior in terms of the KL divergence.

\subsection{Variational estimation in the exponential family}

The joint distribution of observable data and latent variables can be expressed as follows:  
\begin{equation}\label{eq:conditional_conjugacy}
	p(\pmb y_{1:n}, \pmb z_{1:n}, \pmb{\mathbbm P}, \pmb \lambda \vert \alpha_1, \ldots, \alpha_{p-1}, \alpha, \beta, \psi)= p(\pmb{\mathbbm P} \vert \alpha_1, \ldots, \alpha_{p-1}, \alpha, \beta)
	p(\pmb \lambda\vert \psi) 
	\prod_{i=1}^n p(\pmb y_i , \pmb z_i \vert \pmb{\mathbbm P}),
\end{equation}
where the conditional dependencies implicit in \eqref{eq:conditional_conjugacy} highlight the distinction between local latent variables $z_i$ (associated with each observed data point) and global variables $\pmb{\mathbbm P}$ and $\pmb \lambda$, often referred to as parameters. The observable counts and the local latent variables are conditionally independent, given the global parameters.

The original work of \cite{hoffman_stochastic_2013} introduces a general framework for variational inference in exponential families under a hierarchical structure of the form \eqref{eq:conditional_conjugacy}, referred to as conditional conjugacy. This framework relies on the existence of a conditional conjugacy relationship between the prior distribution of the global parameters and the local joint distribution $(\pmb y_i, \pmb z_i)$ for the $i$th observation, where both distributions are assumed to belong to the exponential family (see also \citeauthor{nguyen_depth_2023}, \citeyear{nguyen_depth_2023}). These assumptions are essential for deriving the explicit expression of the optimal global variational parameters that maximize the ELBO. However, the approach proposed in \cite{hoffman_stochastic_2013} is fundamentally based on free-form variational inference, which determines the exact optimal variational distribution by setting the functional derivative to zero, and on the use of the canonical conjugate prior distribution, which, in the case of the Multinomial distribution, corresponds to the Dirichlet distribution.

In order to handle the Beta-Liouville distribution, it becomes necessary to forego the specific Dirichlet-Multinomial conjugation structure. It can be shown that an optimal variational solution can still be achieved by employing fixed-form variational inference, under the assumption that the full conditional distribution of the local variables belongs to an exponential family (fixed hyperparameters are omitted for simplicity):  
\begin{equation}\label{eq:fullconditional1}
	p(\pmb z_i \vert \pmb z_{-i}, \pmb y_{1:n}, \pmb{\mathbbm P}, \pmb \lambda) = h_{\textrm{loc}}(\pmb z_i) \exp\{u_{\textrm{loc}}(\pmb y_i, \pmb{\mathbbm P}, \pmb \lambda )^\top t(\pmb z_i)- a_{\textrm{loc}}\left(u_{\textrm{loc}}(\pmb y_i, \pmb{\mathbbm P}, \pmb \lambda )\right)\},
\end{equation}  
where $\pmb u_{\textrm{loc}} = u_{\textrm{loc}}(\pmb y_i, \pmb{\mathbbm P}, \pmb \lambda )$ represents the natural parameter vector. Furthermore, the local component of the variational distribution is assumed to belong to the same exponential family as $p(\pmb z_i \vert \pmb z_{-i}, \pmb y_{1:n}, \pmb{\mathbbm P}, \pmb \lambda )$:  
\begin{equation}\label{eq:prior1}
	q(\pmb z_i \vert \pmb \gamma_i) = h_{\textrm{loc}}(\pmb z_i) \exp\{\pmb \gamma_i^\top t(\pmb z_i)- a_{\textrm{loc}}\left(\pmb\gamma_i\right)\}.
\end{equation}

Under these more general assumptions, \cite{bilancia_stochastic_2024} demonstrate that the optimal value of the variational hyperparameters in $q(\pmb z_i \vert \pmb \gamma_i)$ that maximize the ELBO (while keeping the other variational hyperparameters fixed) is given by:  
\begin{equation}\label{eq:CAVI1}
	\pmb \gamma_i^\star = \operatorname{E}_{q_V}\left[u_{\textrm{loc}}(\pmb y_i, \pmb{\mathbbm P}, \pmb \lambda ) \right],\quad i=1,\ldots,n,
\end{equation}  
i.e., it corresponds to the expected value, with respect to the variational distribution, of the natural parameters in \eqref{eq:fullconditional1}. Similarly, for the global parameters, assuming hypotheses analogous to those of \eqref{eq:fullconditional1} and \eqref{eq:prior1} (appropriately modified), we derive the optimal values of the respective variational hyperparameters:  
\begin{align}
	\pmb \phi_g^\star & = \operatorname{E}_{q_V} \left[ u_{{\textrm{gl}}, \pmb{\mathbbm P}}(\pmb y_{1:n}, \pmb z_{1:n}) \right], \quad g=1,\ldots, G, \label{eq:CAVI2}\\
	\eta_g^\star & =  \operatorname{E}_{q_V} \left[ u_{{\textrm{gl}}, \pmb \lambda}(\pmb z_{1:n})  \right], \quad g=1,\ldots,G. \label{eq:CAVI3}
\end{align}  

The update equations for the variational parameters \eqref{eq:CAVI1}--\eqref{eq:CAVI3} are derived by computing the expected values with respect to the variational distribution, and they thus depend on the current values of the other variational hyperparameters. A commonly adopted approach is coordinate ascent variational inference (\textsf{CAVI}), where the ELBO is maximized with respect to one parameter at a time, holding the others constant. The estimates of \eqref{eq:CAVI1}--\eqref{eq:CAVI3} are then iteratively updated until convergence is achieved \citep{lee_gibbs_2021}. 

\paragraph{An important caveat} We are now ready to explicitly derive the update equations. However, it is important to emphasize that we have derived all the update equations under the assumption that the parameters appearing were the natural parameters of the representation in exponential form. Nevertheless, this is not always necessarily the case. For instance, if we examine the results presented in the section concerning the local latent variables $\pmb{z}_i$, we notice that the natural parameter is not $\pmb\gamma_i$, but rather $\log(\pmb{\gamma}_i)$ (where the logarithm is taken component-wise). In this case, however, we do not lose generality, as all the calculations remain automatically valid with respect to this new natural parameter, with the only difference being that the update equation for $\pmb{\gamma}_i$ must, of course, be rewritten as (component-wise):
\begin{equation}\label{eq:CAVIalternative}
	\pmb \gamma_i^\star = \exp\left\{\operatorname{E}_{q_V}\left[u_{\textrm{loc}}(\pmb y_i, \pmb{\mathbbm P}, \pmb \lambda ) \right]\right\},\quad i=1,\ldots,n.
\end{equation}  

\paragraph{Global parameter $\pmb{\mathbbm P}$} 

Using the unnormalized posterior distribution \eqref{eq:posterior} and including all terms that do not depend on $\pmb{\mathbbm P}$ in the normalization constant, we obtain the full conditional expression:
\begin{align}
	& p(\pmb{\mathbbm P}\vert \pmb y_{1:n}, \pmb z_{1:n}, \pmb \lambda, \alpha_1, \ldots, \alpha_{p-1}, \alpha, \beta, \psi) \propto
	p(\pmb y_{1:n} \vert \pmb{\mathbbm P}, \pmb z_{1:n}) p(\pmb{\mathbbm P}\vert \alpha_1, \ldots, \alpha_{p-1}, \alpha,\beta) \nonumber \\
	= &  \prod_{i=1}^n p(\pmb y_i \vert \pmb{\mathbbm P}, \pmb z_i) \prod_{g=1}^G p(\pmb\pi_g\vert \alpha_1, \ldots, \alpha_{p-1}, \alpha,\beta), 
\end{align}
from which it follows that:
\begin{align}
 	& \prod_{i=1}^n p(\pmb y_i \vert \pmb{\mathbbm P}, \pmb z_i) \prod_{g=1}^G p(\pmb\pi_g\vert \alpha_1, \ldots, \alpha_{p-1}, \alpha,\beta) \nonumber \\
 	\propto & \prod_{g=1}^G \prod_{\ell=1}^{p-1} \pi_{g\ell}^{\sum_{i=1}^n y_{i\ell} z_{ig}}\left(1 -\sum_{\ell=1}^{p-1} \pi_{g\ell} \right)^{\sum_{i=1}^n y_{ip} z_{ig}}\nonumber \\
 	\times & \prod_{\ell=1}^{p-1}\pi_{g\ell}^{\alpha_\ell-1}\left(\sum_{\ell=1}^{p-1} \pi_{g\ell} \right)^{\alpha-\alpha_0}\left(1 -\sum_{\ell=1}^{p-1} \pi_{g\ell} \right)^{\beta - 1} \nonumber \\
 	= & \prod_{g=1}^G\prod_{\ell=1}^{p-1}\pi_{g\ell}^{\alpha_\ell + \sum_{i=1}^n y_{i\ell} z_{ig} -1}\left(\sum_{\ell=1}^{p-1} \pi_{g\ell}  \right)^{\alpha-\alpha_0}\left(1 -\sum_{\ell=1}^{p-1} \pi_{g\ell} \right)^{\beta + \sum_{i=1}^n y_{ip} z_{ig} - 1},
\end{align}
which is immediately identifiable as the product of $G$ independent $\operatorname{BL}_p$ distributions, with natural parameters:
\begin{align*}
\alpha'_1  & = \alpha_1 + \sum_{i=1}^n y_{i1} z_{ig}, \\
 & \;\;\vdots \\
\alpha'_{p-1} & = \alpha_{p-1} + \sum_{i=1}^n y_{i(p-1)} z_{ig}, \\
\alpha' & = \alpha, \\
\beta' &  = \beta + \sum_{i=1}^n y_{ip} z_{ig}.
\end{align*}

Consequently, the variational distributions are:
\begin{equation}
	q(\pmb \pi_g \vert \pmb\phi_g) = \operatorname{BL}_p(\pmb \pi_g \vert \pmb\phi_g),
\end{equation}
with $\pmb\phi_g = (\phi_{g1},\ldots,\phi_{g(p-1)},\phi_{g\alpha},\phi_{g\beta})^\top$ and independently over $g=1,\ldots,G$. Using the update equation \eqref{eq:CAVI2}, we obtain that the current optimal values are:
\begin{align}
	\phi_{g\ell}^\star & = \alpha_{\ell} + \sum_{i=1}^n y_{i\ell} \gamma_{ig},\quad \ell=1,\ldots,p-1,\\
	\phi_{g\alpha}^\star & = \alpha, \\
	\phi_{g\beta}^\star & = \beta + \sum_{i=1}^n y_{ip} \gamma_{ig}.
\end{align}

It is interesting that $\phi_{g\alpha}^\star$ is not updated, but is fixed to the value of the hyperparameter $\alpha$ set a priori.
$\hfill\blacksquare$

\paragraph{Global parameter $\pmb \lambda$} 
Very similar to the case we just looked at:
\begin{align}
& p(\pmb \lambda\vert \pmb y_{1:n}, \pmb z_{1:n},\pmb{\mathbbm P} , \alpha_1, \ldots, \alpha_{p-1}, \alpha, \beta, \psi) \propto p(\pmb z_{1:n} \vert \pmb \lambda) p(\pmb \lambda \vert\psi) = \prod_{i=1}^n p(\pmb z_i \vert \pmb \lambda)p(\pmb \lambda \vert\psi),
\end{align}
and it is immediate to verify that:
\begin{equation}
\prod_{i=1}^np(z_i \vert \pmb \lambda)p(\pmb\lambda \vert\psi) \propto \prod_{g=1}^G\lambda_g^{\psi + \sum_{i=1}^n z_{ig} - 1},
\end{equation}
is a $G$-dimensional Dirichlet distribution with natural parameters
\begin{equation*}
\psi + \sum_{i=1}^n z_{ig} - 1,\quad g=1,\ldots,G.
\end{equation*}

Therefore, the variational distribution is also a $G$-dimensional Dirichlet distribution:
\begin{equation*}
q(\pmb \lambda \vert \pmb \eta) = \operatorname{Dirichlet}_G(\pmb \lambda \vert \pmb\eta),
\end{equation*}
with $\pmb \eta = (\eta_1, \ldots,\eta_G)^\top$. Using the update equation \eqref{eq:CAVI3} we obtain that:
\begin{equation}
	\eta_g^\star = \psi + \sum_{i=1}^n \gamma_{ig},\quad g=1,\ldots,G.
\end{equation} $\hfill\blacksquare$

\paragraph{Local latent variables $\pmb z_i$}

In this case:
\begin{align}
	& p(\pmb z_i \vert \pmb y_{1:n}, \pmb{\mathbbm P}, \pmb z_{-i}, \pmb \lambda, \alpha_1, \ldots, \alpha_{p-1}, \alpha, \beta, \psi) \propto p(\pmb y_i \vert \pmb z_i , \pmb{\mathbbm P}) p(\pmb z_i \vert \pmb \lambda),
\end{align}
and:
\begin{equation}
	p(\pmb y_i \vert \pmb z_i , \pmb{\mathbbm P}) p(\pmb z_i \vert \pmb \lambda) = p(\pmb y_i, \pmb z_i \vert \pmb{\mathbbm P}, \pmb \lambda) \propto \prod_{g=1}^G \left\{\lambda_g \prod_{\ell=1}^p \pi_{g\ell}^{y_{i\ell}} \right\}^{z_{ig}}.
\end{equation}

Writing this expression in the form of a non-minimal exponential family, we obtain:
\begin{equation}
p(\pmb y_i, \pmb z_i  \vert \pmb{\mathbbm P}) p(\pmb z_i \vert \pmb \lambda) \propto \exp\left\{\sum_{g=1}^G z_{ig}\left[ \sum_{\ell=1}^p y_{i\ell} \log \pi_{g\ell} + \log \lambda_g \right] \right\}
\end{equation}
which is easily recognizable as a $G$-dimensional Multinomial distribution with non-normalized natural parameters:
\begin{equation*}
u_{\textrm{loc}} = u_{\textrm{loc}}(\pmb y_i, \pmb{\mathbbm P}, \pmb \lambda ) \propto \sum_{\ell=1}^p y_{i\ell} \log \pi_{g\ell} + \log \lambda_g, \quad g=1,\ldots,G.
\end{equation*}

The variational distribution is then defined as follows:
\begin{equation*}
	q(\pmb z_i \vert \pmb \gamma_i ) = \operatorname{Multinoulli}_G(\pmb z_i \vert \pmb \gamma_i ),
\end{equation*}
where the natural parameters are $\log\gamma_{ig}$ for $g=1,\ldots,G$. By applying the update equation \eqref{eq:CAVIalternative}, we obtain the following expression for the updated variational parameters:
\begin{equation}\label{eq:normalizedvariational}
	\gamma_{ig}^\star = \frac{\exp \left\{ \sum_{\ell=1}^p y_{i\ell} \operatorname{E}_{q_V} \left[  \log \pi_{g\ell} \right] + \operatorname{E}_{q_V} \left[ \log \lambda_g \right] \right\}}{\sum_{g=1}^G \exp \left\{ \sum_{\ell=1}^p y_{i\ell} \operatorname{E}_{q_V} \left[  \log \pi_{g\ell} \right] + \operatorname{E}_{q_V} \left[ \log \lambda_g \right] \right\}}.
\end{equation}

Utilizing the standard properties of exponential families and the Dirichlet form of the variational distribution for $\pmb \lambda$, we obtain:
\begin{equation}
	\operatorname{E}_{q_V} \left[ \log \lambda_g \right] = \operatorname{E}_{q_{\pmb \lambda}} \left[ \log \lambda_g \right] = \Psi(\eta_g) - \Psi\left( \sum_{g=1}^G \eta_g \right).
\end{equation}

Similarly, from \eqref{eq:blsufficient2}, we have:
\begin{equation}
	\operatorname{E}_{q_{\pmb \pi_g}}\left[\log \pi_{g\ell} \right] =
	\Psi(\phi_{g\ell}) - \Psi\left(\sum_{\ell = 1}^{p-1} \phi_{g\ell}\right) + \Psi(\alpha) - \Psi(\alpha + \phi_{g\beta}), \quad \ell=1,\ldots,p-1,
\end{equation}
and from \eqref{eq:blsufficient1}, we obtain:
\begin{equation}
	\operatorname{E}_{q_{\pmb \pi_g}}\left( \log\left[1 - \sum_{\ell = 1}^{p-1} \pi_{g\ell} \right] \right) = \Psi(\phi_{g\beta}) - \Psi(\alpha + \phi_{g\beta}).
\end{equation}

These expression can be substituted into \eqref{eq:normalizedvariational} using the current estimates of the variational parameters.

\subsection{Stochastic variational inference}

As previously discussed in Subsection \ref{sec:dirmultisec}, we introduce a stochastic variant of the proposed variational algorithm, where the expected values in \eqref{eq:CAVI2} and \eqref{eq:CAVI3}, which are computed over the entire dataset, are substituted with a noisy but unbiased and computationally inexpensive estimate derived from a single sampled data point $\pmb z_s$ \citep{hoffman_stochastic_2013}, with $s \sim \mathsf{Uniform} \left(1,2,\ldots,n\right)$. The algorithm leverages this sampled data point at each iteration, and the update equation for the local latent variables at iteration $t$ is given (for $g=1,\ldots,G$) by:
\begin{equation}
\widetilde\gamma_{sg}^{(t)} \propto \exp \left\{ \sum_{\ell=1}^p y_{s\ell}\operatorname{E}_{q_V} \left[  \log \pi_{g\ell} \right] + \operatorname{E}_{q_V} \left[ \log \lambda_g \right] \right\},
\end{equation}
distinguish between $\widetilde{\gamma}_{sg}$, which is computed for a single data point, and $\gamma_{ig}^\star$, which must be calculated for each data point in every cycle, totaling $n$ evaluations. The global parameters are temporarily set to the following intermediate values (for $g=1,\ldots,G$):
\begin{align}
	\widehat\phi_{g\ell}^{(t)} & = \alpha_{\ell} + n y_{s\ell} \widetilde\gamma_{sg}^{(t)},\quad \ell=1,\ldots,p-1,\\
	\widehat\phi_{g\beta}^{(t)} & = \beta + n y_{sp} \widetilde\gamma_{sg}^{(t)}\\
	\widehat\eta_g^{(t)} &= \psi + n \widetilde\gamma_{sg}^{(t)},	
\end{align}
and then updated using an equation with an adaptive step size (for simplicity, we provide only the first equation, as the others follow an identical structure):
\begin{equation}\label{eq:ewma}
	\widetilde\phi_{g\ell}^{(t+1)} = (1 - \rho^{(t)}) \widetilde \phi_{g\ell}^{(t)}  + \rho^{(t)}  \widehat\phi_{g\ell}^{(t)}.
\end{equation}

The adaptive step size $\rho^{(t)}$ must satisfy the Robbins-Monro conditions to ensure convergence \citep{satoOnlineModelSelection2001,kushnerStochasticApproximationRecursive2003,blei_variational_2017,bottouOptimizationMethodsLargeScale2018}:
\begin{equation}\label{eq:robbinsmonro}
	\sum_{t=1}^{\infty} \rho^{(t)} = \infty,\quad \sum_{t=1}^{\infty} \left(\rho^{(t)}\right)^2 < \infty.
\end{equation}

For instance, one might choose $\rho^{(t)} = (1 + t)^{-\kappa}$, with $\kappa \in (0.5,1]$ \citep{tranPracticalTutorialVariational2021a}. The forgetting rate $\kappa$ controls the speed at which past information is downweighted in the exponentially weighted moving average \eqref{eq:ewma}. A higher value of $\kappa$ (closer to 1) results in greater downweighting of earlier values. The entire computational flow is outlined in Algorithm \ref{alg:algo1}, and it can be demonstrated that this procedure corresponds to a gradient ascent method, where the local parameters are updated with a unit step size, while the global parameters are updated using adaptive steps of size $\rho^{(t)}$. Consequently, this approach is equivalent to minimizing $-\!\operatorname{ELBO}(q_V)$.

\begin{algorithm}[!ht]
	\caption{Stochastic version of the \textsf{CAVI} algorithm (\textsf{SVI})}\label{alg:algo1}
	\begin{algorithmic}[1]
		\Require Data $\pmb y_{1:n}=(\pmb y_1,\pmb y_2,\ldots,\pmb y_n)$; number of components $G$; prior hyperparameters $\alpha_1,\alpha_2,\ldots,\alpha_{p-1},\alpha,\beta,\psi > 0$; variational families $q(\pmb z_i \vert \pmb \gamma_i)$, $q(\pmb \pi_g \vert \pmb\phi_g)$ for $g=1,\ldots,G$, $q(\pmb \lambda \vert \pmb \eta)$; step size $\rho^{(t)}$ for $t=1,2,\ldots$
		\INPUT Variational parameters $\gamma_{ig}$ for $i=1,\ldots,n$, $g=1,\ldots,G$; $\phi_{g\ell}$ for $g=1,\ldots,G$, $\ell=1\ldots, (p+1)$; $\eta_g$ for $g=1,\ldots,G$ (initialize randomly in $t=0$).
		\Ensure Optimized variational densities.  
		\While{the {\textsf ELBO} has not converged, for $t=1,2,\ldots $}:
		\State Sample $s \sim \mathsf{Uniform}(1,2,\ldots,n)$
		\State Update local variational parameters:
		\For{$g=1,\ldots,G$}
		\State $\gamma_{sg} \gets \exp \left\{ \sum_{\ell=1}^p y_{s\ell}\operatorname{E}_{q_V} \left[  \log \pi_{g\ell} \right] + \operatorname{E}_{q_V} \left[ \log \lambda_g \right] \right\}$
		\State $\gamma_{sg} \gets \frac{\gamma_{sg}}{\sum_{h=1}^G \gamma_{sh}}$ 
		\EndFor
		\State Intermediate global variational parameters:
		\For{$g=1,\ldots,G$}
		\For{$\ell=1,\ldots, p-1$}
		\State $\widehat\phi_{g\ell} \gets \alpha_{\ell} + n y_{s\ell} \gamma_{sg}$
		\EndFor
		\State $\widehat\phi_{g\beta} \gets \beta + n y_{sp} \gamma_{sg}$
		\EndFor
		\For{$g=1,\ldots,G$}
		\State $\widehat\eta_g \gets \psi + n \gamma_{sg}$
		\EndFor
		\State Update global variational parameters:
		\For{$g=1,\ldots,G$}
		\For{$\ell=1,\ldots, p-1$}
		\State $\phi_{g\ell} \gets \left(1 - \rho^{(t)}\right) \phi_{g\ell} + \rho^{(t)}\widehat \phi_{g\ell}$
		\EndFor
		\State $\phi_{g\beta} \gets \left(1 - \rho^{(t)}\right) \phi_{g\beta} + \rho^{(t)}\widehat \phi_{g\beta}$
		\EndFor
		\For{$g=1,\ldots,G$}
		\State $\eta_{g}\gets \left(1 - \rho^{(t)}\right) \eta_{g} + \rho^{(t)}\widehat \eta_{g}$
		\EndFor
		\EndWhile
	\end{algorithmic}
\end{algorithm}

\section{Data examples}\label{sec:dataexamples}

In the following examples, the definition of the hyperparameters for the Dirichlet-Multinomial model described in Subsection \ref{sec:dirmultisec} is based on the assumption (commonly used in real-world case studies) that no prior information is available. Consequently, we selected $\theta= 1$ and $\psi = 5/G$, a configuration that remains neutral with respect to both $\theta$ and $\psi$. The hyperparameters of the Beta-Liouville Multinomial model were determined following the same rationale, with:
\begin{align}
	\alpha_\ell & = 1, \quad \ell=1,\ldots,p-1,\\
	\beta & = 1.
\end{align}

From this, it immediately follows that $\alpha_0 = p-1$. Since our objective is to compare the Dirichlet prior distribution with the Beta-Liouville, the parameter $\alpha$ is selected to introduce an increasing divergence from the Dirichlet distribution, as defined by:  
\begin{equation}\label{eq:alphasetting}
	\alpha = \alpha_0 + \delta \alpha_0,
\end{equation}  
where $\delta = 0, \pm 0.05, \pm 0.10, \pm 0.20, \pm 0.30, \pm 0.40, \pm 0.50$. Based on the results presented in Subsection \ref{sec:betaliouvilledef}, the Dirichlet distribution is recovered at $\delta = 0$, while the divergence increases with the absolute value of $\delta$. All simulations were conducted with $n=\num[group-separator={,}]{5000}$ iterations of the main cycle described in Algorithm \ref{alg:algo1}. Clearly, this choice is suboptimal, as the convergence of the ELBO could be diagnosed before reaching $n=5000$. Nevertheless, it enables comparisons between the different settings in terms of the CPU time required for a single execution of the algorithm.

In general, the convergence of the \textsf{CAVI} algorithm can be assessed visually, as the ELBO is guaranteed to increase monotonically across iterations. Any deviation from this behavior signals a programming error in the code. Naturally, utilizing the ELBO as a convergence criterion imposes a significant additional computational cost on each iteration. Based on the expression in the Appendix \ref{sec:A1}, this cost is at least of the order $O(n)$. Nevertheless, this form of monitoring is crucial to ensure that the algorithm has reached a mode of the surface that approximates the posterior distribution of the parameters. It is equally important to emphasize that, in the case of \textsf{SVI}, the ELBO may exhibit local decreases due to the algorithm's inherent randomness, although global convergence is guaranteed under the Robbins-Monro conditions \eqref{eq:robbinsmonro}. However, the ELBO is typically a non-concave function, implying that \textsf{SVI} ensures convergence only to a local optimum, which may be influenced by the choice of initial values \citep{blei_variational_2017,plummer_dynamics_2020}. In other words, variational inference explores a single mode of the posterior surface, and each execution of the algorithm converges to a distinct local maximum of the ELBO. Consequently, multiple runs with randomly varying initializations are required to identify the optimal solution. To address this, we executed the \textsf{SVI} algorithm 30 times for each hyperparameter configuration, employing pseudorandom generators to initialize the starting values while respecting the constraints imposed by the parameters. For example, a Gamma distribution with large variance was utilized to generate the initial values of $(\phi_{g1},\ldots,\phi_{g(p-1)},\phi_{g\beta})^\top$.

The assignment to clusters with a fixed number of groups $G$ was performed using the standard allocation function:  
\begin{equation}\label{eq:argmaxclass}
	g_i^{\mathsf{ MAP }}=\argmaxF_{g=1,\ldots, G} \operatorname{E}_{q_{\pmb\gamma}}(z_{ig}\vert \gamma_{i}^\star) = \argmaxF_{g=1,\ldots, G}\gamma_{ig}^\star,
\end{equation}  
which is known to minimize the expected posterior loss under a 0/1 loss function, penalizing incorrect assignments \citep{james2023introduction}. Although clustering is an unsupervised algorithm, in our case the ground truth is available, enabling the evaluation of clustering quality using the Adjusted Rand Index (ARI) as well as an external measure of accuracy. The latter is based on the best match between the true labels $g_i$ and the estimated labels $\hat g_i \equiv g_i^{\mathsf{ MAP }}$, defined as follows:  
\begin{equation}\label{eq:permacc}
	\operatorname{accuracy} = \max_{\mathbbm p \in \mathcal P} \frac{1}{n} \sum_{i=1}^{n} \mathbbm 1\left(g_i = \mathbbm p(\hat{g}_i)\right),
\end{equation}  
where $\mathcal P$ is the set of all permutations in $\left\{1,\ldots,G\right\}$. To solve the optimization problem in (\ref{eq:permacc}), any implementation of the Hungarian algorithm with polynomial time complexity $O(G^3)$ can be used \citep{hungarian2023}. The \textsf{SVI} algorithm was implemented in \textsf{R} 4.4.1, along with several additional libraries to support the coding process \citep{RCRAN}.  

\subsection{Reuters-21578}\label{subsec:reuterse21578acc}

The dataset is a subset of the \textsf{Reuters 21578} collection \citep{apteAutomatedLearningDecision1994,lewisRCV1NewBenchmark2004}, obtained by sampling 10\% of the documents within the $G=5$ categories listed in the first column of Table \ref{tab:tab_reutersmulti}. To convert the documents into a bag-of-words (BOW) representation, the raw text data was preprocessed using \textsf{R} 4.4.1 and the \texttt{tm} library \citep{tmlibrary} through the following steps: removal of extra spaces, punctuation marks, and numbers; conversion to lowercase; elimination of stop words; stemming to reduce inflectional forms to their common base form; recompletion using the most frequent match; and tokenization into unigrams, retaining only tokens between 4 and 16 characters in length and discarding the rest. 
During preprocessing, all sparse terms with a document-level frequency below 1\% were removed from the term vocabulary. The resulting document-term matrix had dimensions $(n = 750) \times (p = 754)$, with an overall sparsity of 96\% and an average of 27.55 terms per document, indicating that the texts were very short.

\begin{table*}[!ht]
	\centering
	\caption{The subset from \textsf{Reuters 21578} used in the example, obtained by sampling with a sampling fraction of 10\% of the $G=5$ categories shown in the first column.}
	\label{tab:tab_reutersmulti}
	\vskip 0.1cm
	\begin{tabular}{ccl} 
		category &  \multicolumn{1}{c}{nr.~of docs} & \multicolumn{1}{c}{topic}  \\ \hline\hline
		\textsf{acq}       & 221  & corporate mergers and acquisition   \\
		\textsf{crude}   & 50 & crude oil price  \\
		\textsf{earn}     & 375  & earning reports \\
		\textsf{grain}    & 44  & grain market and trade \\
		\textsf{money-fx}  & 60 & foreign exchange market \\ \hline
		total                         & $n=750$ &\\
		\hline\hline
	\end{tabular}
\end{table*}

\begin{figure*}[!h] 
	\begin{center}
		\includegraphics[width=0.49\linewidth]{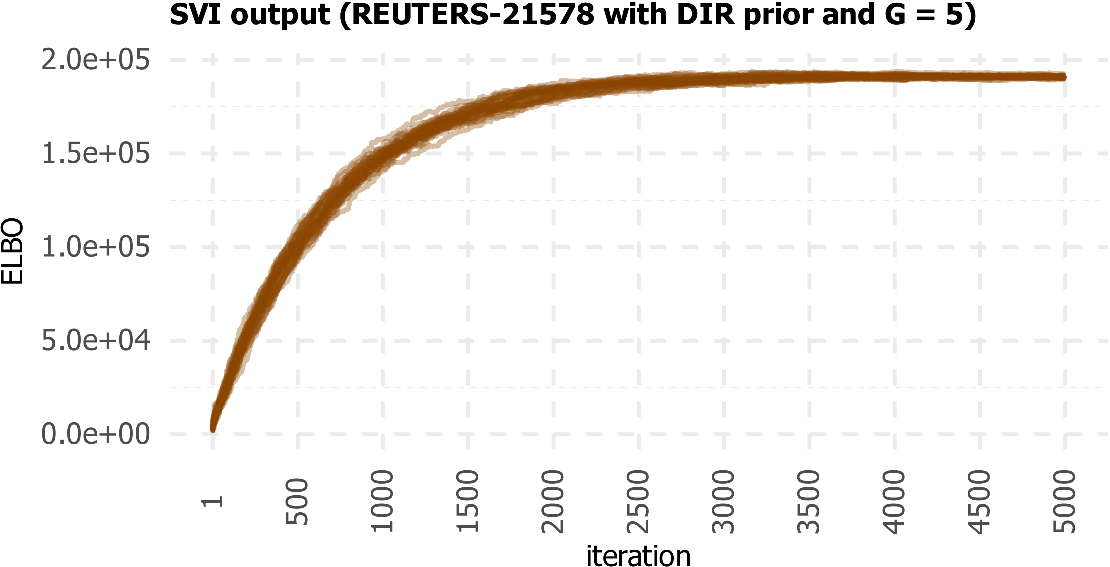}
		\includegraphics[width=0.49\linewidth]{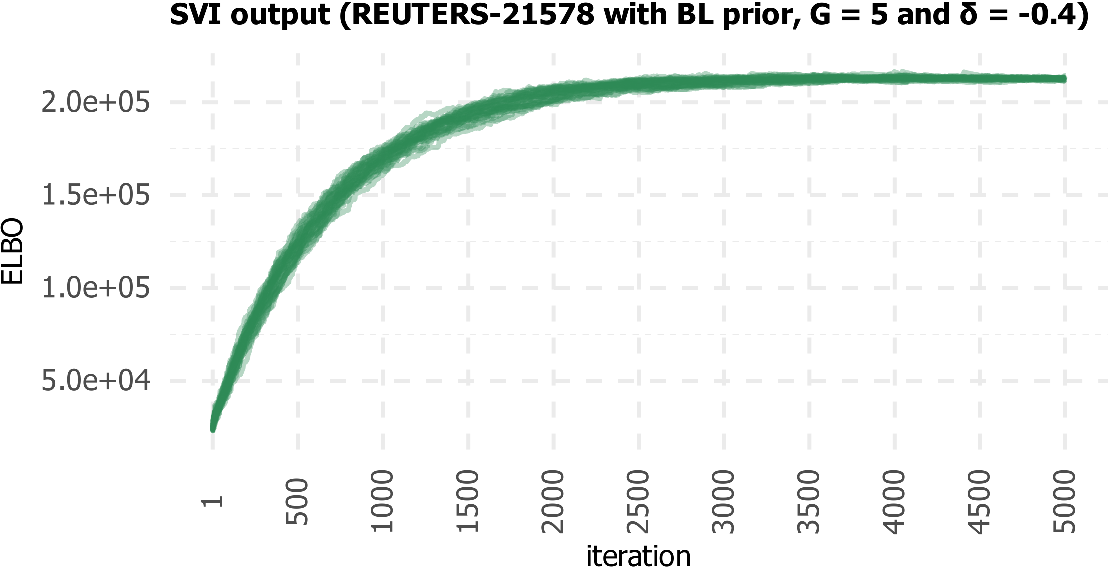}
	\end{center}     
	\caption{Left: 30 runs of the \textsf{SVI} algorithm on a subset of \textsf{Reuters 21578} with $G = 5$ categories, using the Dirichlet prior (DIR) and a total of \num[group-separator={,}]{5000} iterations. Right: same as the left panel, except that a Beta-Liouville prior with $\delta = -0.4$ (BL) was used. In both cases, a forgetting rate of $\kappa = 0.6$ was applied.}\label{fig:elbo_figs_reuters}
\end{figure*}

Figure \ref{fig:elbo_figs_reuters} presents the trajectories of 30 runs of the \textsf{SVI} algorithm for two priors: the Dirichlet prior (left) and the Beta-Liouville prior (right) with $\delta = -0.4$ (this value of $\delta$ is used as an example and holds no special significance). All simulations were conducted with a forgetting rate of $\kappa = 0.6$, although the results remain essentially invariant for $\kappa \in (0.5, 0.9]$.

\begin{figure*}[!h] 
	\begin{center}
		\includegraphics[width=0.49\linewidth]{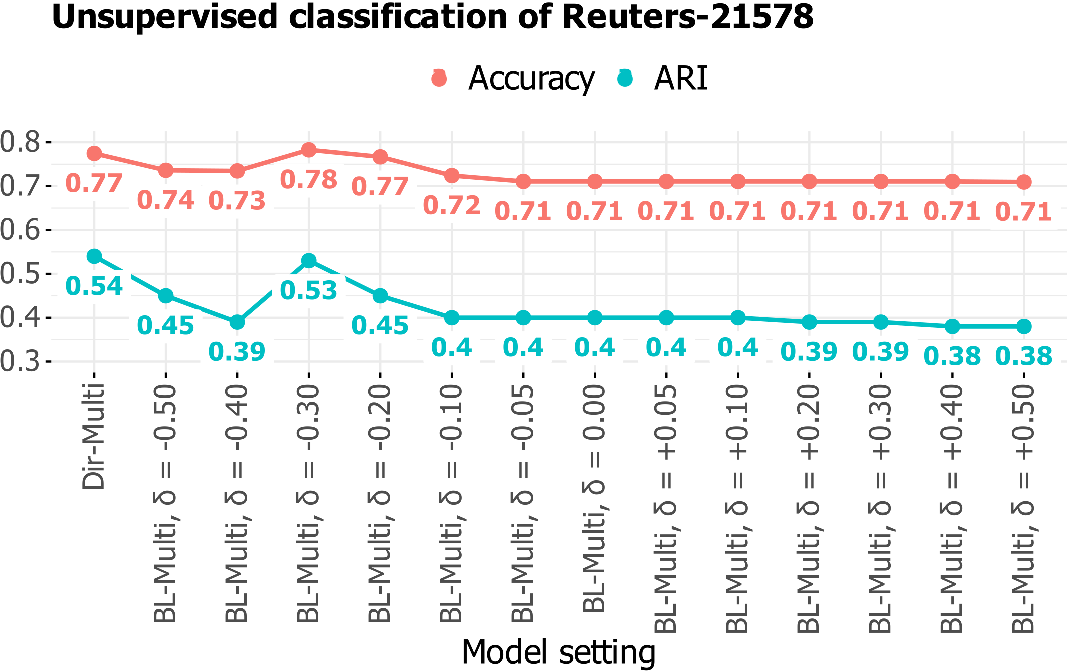}
		\includegraphics[width=0.49\linewidth]{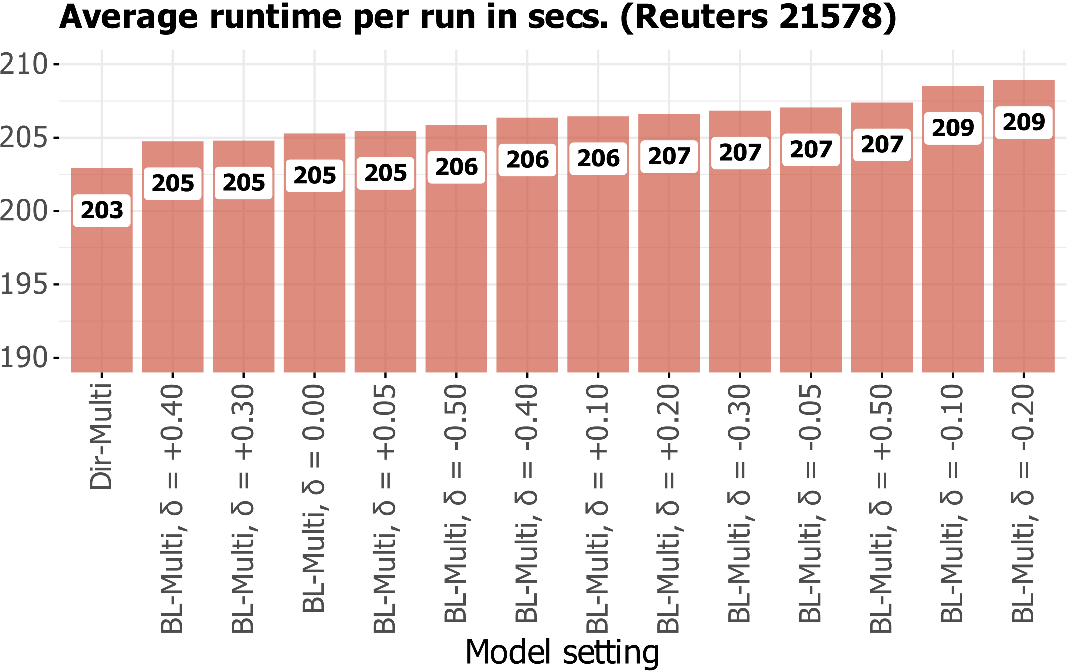}
	\end{center}     
	\caption{Left: Permutation-based accuracy \eqref{eq:permacc} and Adjusted Rand Index (ARI) for clustering the \textsf{Reuters-21578} corpus, where documents are assigned to clusters according to the allocation function \eqref{eq:argmaxclass}, which corresponds to the 0/1 posterior loss. Right: Average runtime (over 30 runs) in seconds for a single run of Algorithm \ref{alg:algo1}.}\label{fig:accuracy_figs_reuters}
\end{figure*}

The results for clustering accuracy, based on the metrics considered, are presented in Figure \ref{fig:accuracy_figs_reuters} (left panel). It is evident that the use of a Beta-Liouville prior does not result in significantly worse clustering accuracy compared to the Dirichlet prior. For $\delta = -0.3$, the highest accuracy achieved is 0.78, slightly higher than the 0.77 achieved with the Dirichlet distribution. The best ARIs are also comparable (Dirichlet: 0.54 vs. Beta-Liouville: 0.53). Regarding the average run times for a single run of stochastic variational inference, the right panel of Figure \ref{fig:accuracy_figs_reuters} shows no significant differences, with the observed deviations likely attributable to random variations.
Additionally, when considering the \textsf{SVI} algorithm described in \cite{bilancia_stochastic_2024} for the Dirichlet-Multinomial case, it becomes apparent that, for the Beta-Liouville prior, the number of update equations at each iteration remains identical. Any systematic differences (though no conclusive tests were performed) are likely due to the increased computational complexity involved in evaluating the expressions \eqref{eq:blsufficient1}-\eqref{eq:blsufficient2}.

Two more interesting considerations arise. First, the accuracy obtained with $\delta = 0$ (0.71) is lower than that achieved with the Dirichlet-Multinomial model. While this may seem surprising, it is actually due to the different geometries of the spaces in which the ELBO is maximized. We will revisit this point in the next section. The second, even more intriguing observation is that the accuracy of the clustering obtained with the Beta-Liouville prior decreases as $\delta$ becomes positive and, in any case, is inferior to the accuracy achieved with the Dirichlet-Multinomial model when $\delta > +_{}0.10$.

\begin{figure*}[!h] 
	\begin{center}
		\includegraphics[width=0.75\linewidth]{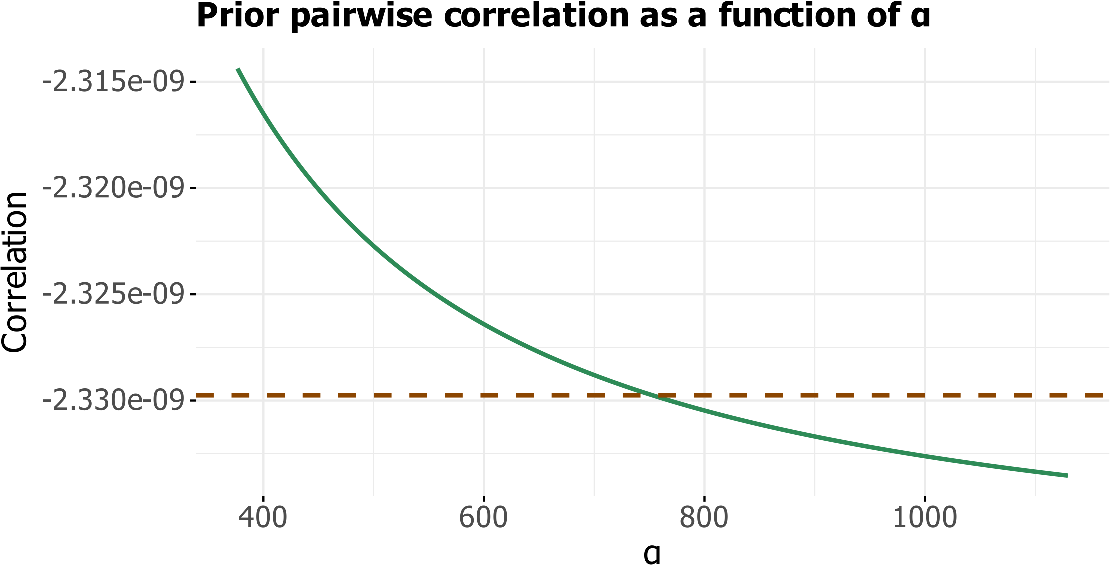}
	\end{center}     
	\caption{Solid green line: Correlation function \eqref{eq:eqcorrbl} between word pairs, where $\delta$ varies between -0.5 and +0.5 in steps of 0.001 and $\alpha$ is selected according to \eqref{eq:alphasetting}. Dashed brown line: Correlation between word pairs with the Dirichlet prior.}\label{fig:corr_fig}
\end{figure*}

To provide an interpretation, we have plotted the correlation function \eqref{eq:eqcorrbl} between pairs of words in Figure \ref{fig:corr_fig}. In this plot, $\delta$ varies between -0.5 and +0.5 in increments of 0.001, and $\alpha$ is set according to \eqref{eq:alphasetting} (solid green line). The dashed horizontal line represents the correlation under the Dirichlet prior. Although these prior correlations are updated in the posterior, it is clear that optimal performance is achieved within the range of $\alpha$ values corresponding to negative $\delta$, where the negative pairwise correlation is weaker than under the Dirichlet prior. 
In the scientific literature, the term `short text' lacks a consistent definition. Intuitively, the limited context and sparsity of short texts tend to induce negative correlations in frequency-based representations. However, the repulsive behavior of the Dirichlet distribution seems overly extreme in this context, suggesting that a Beta-Liouville prior with $\alpha$ within an appropriate range may more accurately model such texts.

\subsection{BBCSport}

The BBCSport dataset forms part of a comprehensive benchmark collection commonly employed in text mining research \citep{greenePracticalSolutionsProblem2006}. The corpus examined in this study consists of $n=737$ documents retrieved from the BBC Sport website, encompassing news articles on five sports spanning the years 2004 to 2005:
\begin{itemize}
	\item[--] \texttt{athletics} (101 documents, 13.70\%)
	\item[--] \texttt{cricket} (124 documents, 16.82\%)
	\item[--] \texttt{football} (265 documents, 35.96\%)
	\item[--] \texttt{rugby} (147 documents, 19.95\%)
	\item[--] \texttt{tennis} (100 documents, 13.57\%)
\end{itemize}

The documents within the five classes exhibit a notable degree of semantic overlap. While they pertain to different sports, they share a common thematic focus characteristic of content generated by a sports newsroom. The corpus underwent the same pre-processing steps as in the previous example, yielding a large matrix with dimensions $737 \times 7883$ and a high sparsity level of 98.58\%. Following the removal of terms appearing in fewer than 5\% of the total documents, the final document-term matrix was $(n = 737) \times (p = 506)$, with an overall sparsity of 88\% and an average of approximately 59 words per document.

\begin{figure*}[!h] 
	\begin{center}
		\includegraphics[width=0.49\linewidth]{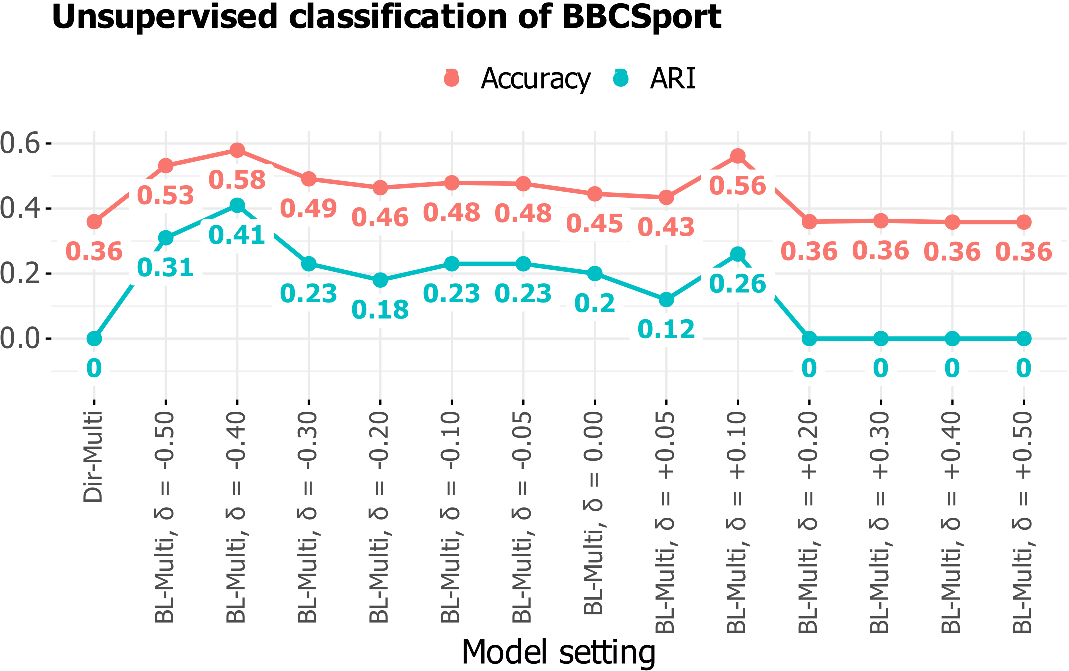}
		\includegraphics[width=0.49\linewidth]{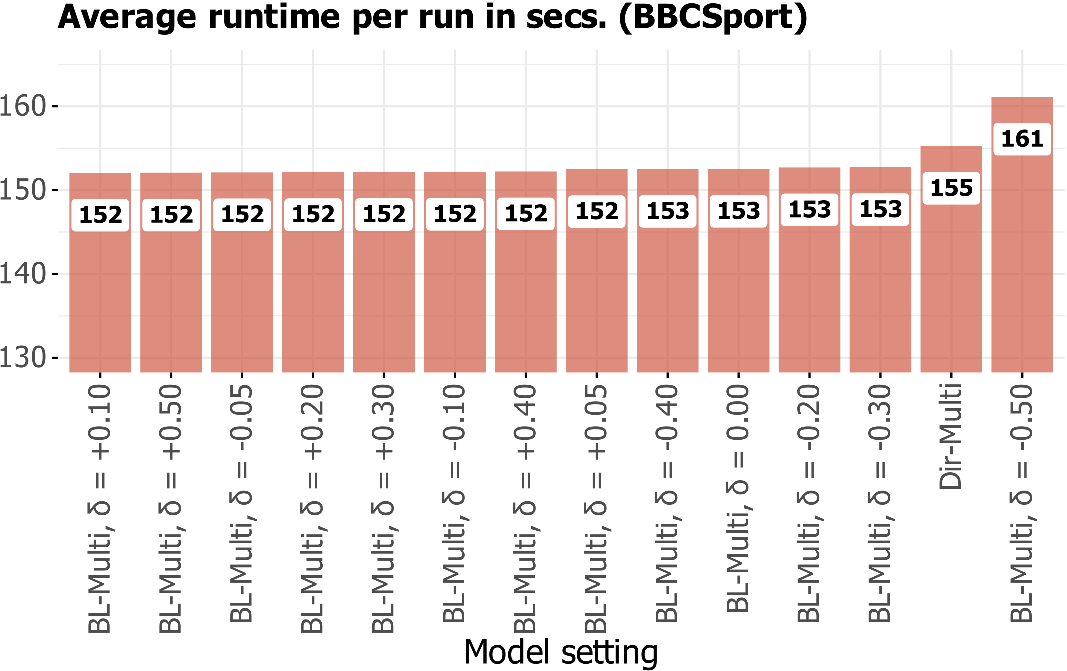}
	\end{center}     
	\caption{Left: Permutation-based accuracy \eqref{eq:permacc} and Adjusted Rand Index (ARI) for clustering the BBCSport corpus, where documents are assigned to clusters under the allocation function \eqref{eq:argmaxclass} that applies to the 0/1 posterior loss. Right: Average runtime (over 30 runs) in seconds for a single run of the Algorithm \ref{alg:algo1}.}\label{fig:accuracy_figs_bbcsports}
\end{figure*}

The results are presented in Figure \ref{fig:accuracy_figs_bbcsports}. From the perspective of the computation time required for a single run (right panel), it is largely confirmed that the observed differences between the two prior distributions can be attributed to random variations. Interestingly, the results depicted in the left panel of the same figure reveal that the Dirichlet-Multinomial model fails to capture any discernible structure within this textual corpus, likely due to the aforementioned semantic ambiguity. Conversely, the Beta-Liouville Multinomial model demonstrates significantly better performance, particularly when $\delta$ falls within the same range of negative values as in the previous example. Specifically, with $\delta = -0.40$, the model achieves the highest accuracy (0.58) along with an ARI value (0.41), indicating a substantial departure from a purely random partition. Comparable accuracy and ARI values are observed at $\delta = +0.10$, whereas performance declines when $\delta > +0.10$, reverting to levels comparable to those of the Dirichlet-Multinomial model.

\subsection{Topic quality (Reuters-21578)}

Clustering accuracy is not the sole criterion for evaluating the proposed model. An equally important aspect is the semantic coherence of the probability distributions estimated over the vocabulary $\mathbbm{V}$. These distributions define low-dimensional subspaces (topics), each of which can be interpreted as conveying specific semantic content. Although this construct is theoretically plausible, text categorization models often generate low-dimensional subspaces that lack interpretability for human domain experts. In other words, coherence refers to the degree to which the words within a topic are semantically related to one another and how effectively they capture the overall theme of the topic \citep{newman2010,Roder2015}.

Using once again the Reuters-21578 dataset, pre-processed as described in Subsection \ref{subsec:reuterse21578acc}, we computed, for each possible value of the hyperparameter $\delta$, the following coherence metric \citep{mimno_optimizing_2011,meaney_quality_2023}:
\begin{equation}\label{eq:topiccoherence}
	\mathcal C(\mathbbm V^{(\mathbbm t, M)}) = \sum_{m=2}^M \sum_{s=1}^{m-1}\log\frac{ D(v_m^{(\mathbbm t)}, v_s^{(\mathbbm t)}) + 1 }{D(v_s^{(\mathbbm t)})},
\end{equation}
where $\mathbbm{V}^{(\mathbbm t,M)} = (v_1^{(\mathbbm{t})}, v_2^{(\mathbbm{t})}, \ldots, v_M^{(\mathbbm{t})})$ denotes the list of the $M$ most probable terms for topic $\mathbbm{t}$, and $D(v, v')$ represents the co-document frequency of terms $v$ and $v'$--that is, the number of documents in which both terms appear at least once. Based on expert annotations, \citep{mimno_optimizing_2011} demonstrate the effectiveness of $\mathcal C(\mathbbm V^{(\mathbbm t, M)})$, as defined in \eqref{eq:topiccoherence}, in identifying poorly formed topics, with lower values indicating lower semantic quality.

\begin{figure*}[!h] 
	\begin{center}
		\includegraphics[width=0.75\linewidth]{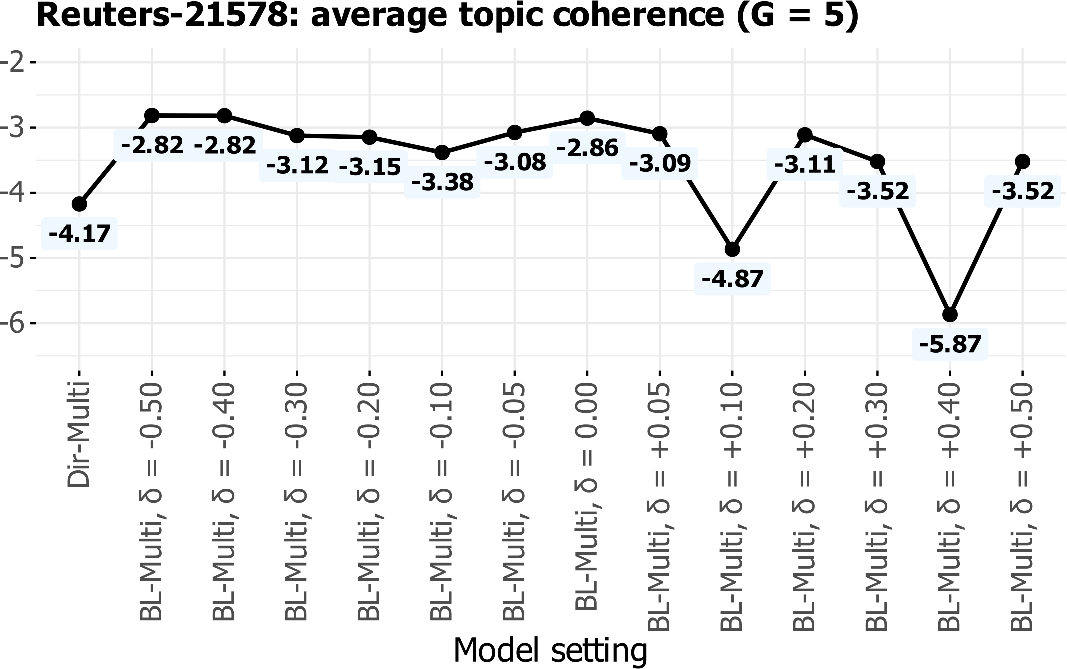}
	\end{center}     
	\caption{Average topic coherence $\mathcal C(\mathbbm V^{(\mathbbm t, M)})$ as defined in equation \eqref{eq:topiccoherence} with $M = 10$, averaged over the $G=5$ topics estimated for each model.}\label{fig:coherence_reuters21578}
\end{figure*}

The results are presented in Figure \ref{fig:coherence_reuters21578}. Since a distinct coherence score is computed for each topic, we report the average coherence across the $G = 5$ topics, comparing the model based on the Dirichlet-Multinomial prior with the Beta-Liouville Multinomial model. Once again, the Dirichlet-Multinomial model demonstrates substantially poorer performance relative to the Beta-Liouville Multinomial model when $\delta = 0$. This further highlights the phenomenon whereby, despite the equivalence of the underlying generative models, the variational inference algorithm explores fundamentally different solution spaces. Furthermore, in line with the classification accuracy results, the most robust performance is achieved by the Beta-Liouville Multinomial model for negative values of $\delta$, with peak coherence observed at $\delta = -0.40$ or $\delta = -0.30$. For positive values of $\delta$, performance declines, in some cases even falling below that of the Dirichlet-Multinomial model.

\begin{figure*}[!h] 
	\begin{center}
		\includegraphics[width=0.49\linewidth]{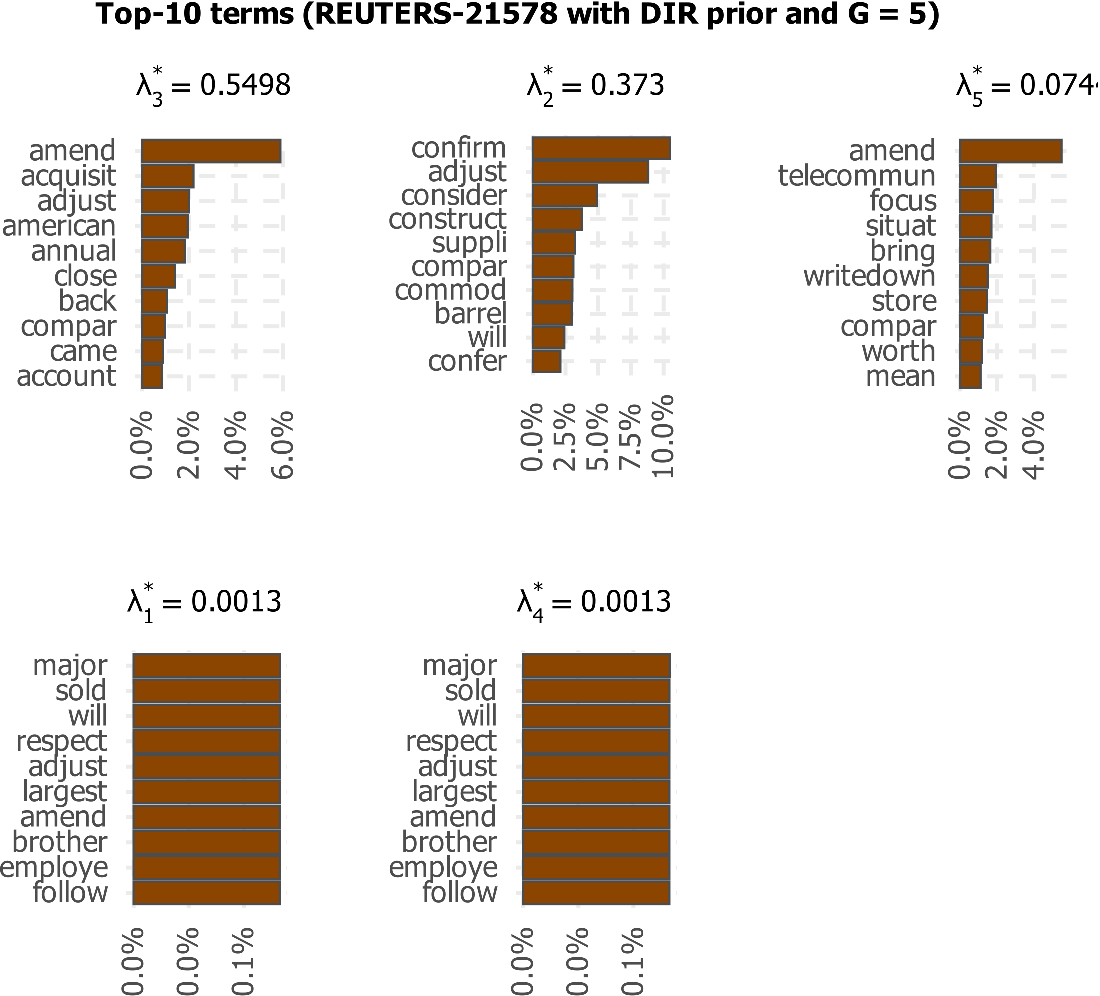}
		\includegraphics[width=0.49\linewidth]{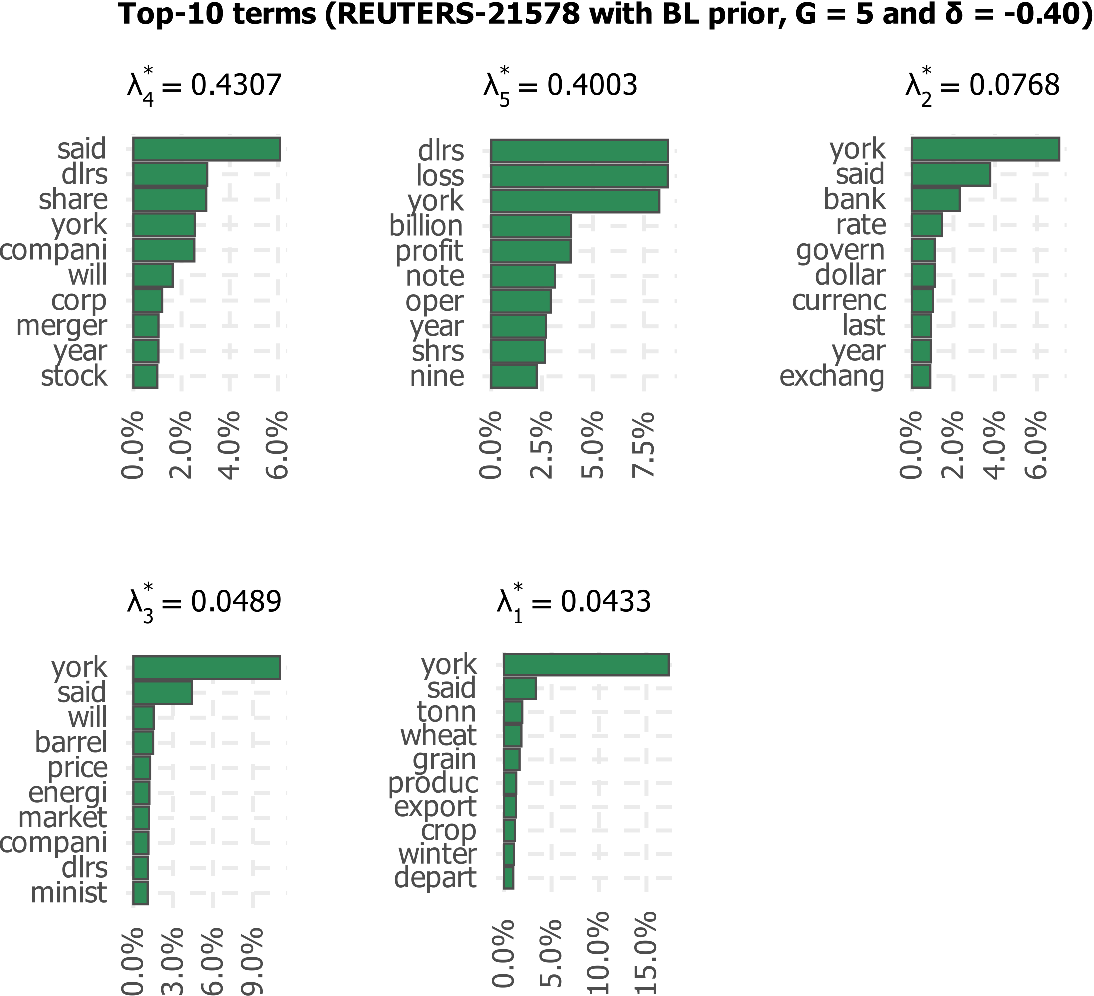}
	\end{center}     
	\caption{Left: Top-10 terms from the \textsf{SVI}-based estimates of each row of the $\mathbbm{P}$ matrix in the Dirichlet-Multinomial model. The rows of the estimated matrix $\mathbbm{P}^\star$ were ordered according to the estimated component weights $\lambda_g^\star$ ($g = 1, 2, \ldots, G$). The index assigned to each component is purely conventional, as there are $G!$ equivalent modes that differ only by permutation. Right: Top-10 terms from the \textsf{SVI}-based estimates of each row of the $\mathbbm{P}$ matrix in the Beta-Liouville Multinomial model, with $\delta = -0.40$.}\label{fig:topdrivers}
\end{figure*}

Provided that the variational estimates can be used to approximate the posterior estimates of the probability distributions $\pmb{\pi}_g$ on $\mathbbm{V}$ as follows:
\begin{equation}\label{eq:beta_est}
	\pi_{g\ell}^\star = \frac{\phi_{g\ell}^\star}{\sum_{s=1}^p \phi_{gs}^\star}, \quad g = 1, 2, \ldots, G, \quad \ell = 1, 2, \ldots, p-1,
\end{equation}
we have employed these estimates to identify the 10 most salient words in terms of their probability of occurrence. These probabilities are depicted in Figure \ref{fig:topdrivers} for each of the two models, utilizing the choice $\delta = -0.40$, which guarantees peak performance for the Beta-Liouville Multinomial model. Each set of distributions is sorted based on the estimated mixing weights $\lambda_j^\star$ of each component (in descending order). The index assigned to each component is purely conventional, given the existence of $G!$ equivalent modes differing solely by a permutation of the indices (although variational algorithms explore only one mode at a time).

With the Dirichlet-Multinomial model, it becomes apparent that the resulting solution suffers from weak identifiability, as two identical components are perfectly overlapping--at least in terms of their top-10 words. This constitutes a form of non-identifiability that departs from the classical label switching invariance typically associated with finite mixture models. In essence, the Dirichlet-Multinomial model effectively recovers a mixture with $G = 3$ distinct components, along with two redundant, superimposed ones. Furthermore, establishing a meaningful correspondence with any of the original topics, as reported in Table \ref{tab:tab_reutersmulti}, proves particularly difficult. Consequently, although the clustering accuracy is substantially better than that of a purely random assignment, the dimensionality reduction achieved by the Dirichlet-Multinomial model remains unsatisfactory, as the estimated topics do not support meaningful unsupervised interpretation by a human user.

With the Beta-Liouville Multinomial model ($\delta = -0.40$), the situation changes markedly. Notably, identifiability issues are absent, and the weights of the five components approximately reflect the relative frequencies of the $G = 5$ categories reported in Table \ref{tab:tab_reutersmulti}. In the first distribution, the presence of terms such as \texttt{compani}, \texttt{stock}, and \texttt{merger} (stemmed terms have not been recompleted) clearly indicates a strong correspondence with the \textsf{acq} category. Likewise, for the remaining topics, a cursory analysis reveals a reasonably clear alignment between the top-10 words and the categories of the Reuters-21578 subset employed. For instance, the final distribution, which accounts for 4.33\% of the weight--corresponding to approximately 32 documents out of the original 44--exhibits a clear association with the \textsf{grain} category. The regularization effect induced by the Beta-Liouville prior is especially evident in this case, particularly when contrasted with the outcomes produced by the standard Dirichlet-Multinomial model.

\section{Discussion and conclusions}\label{sec:discussion_conclusions}

The Unigram mixture model has become a standard tool for analyzing and performing unsupervised classification on text corpora composed of short documents, such as abstracts, emails, social media posts, and web news. From a Bayesian perspective, inference is often performed using the Dirichlet distribution as a prior over the term vocabulary. In this paper, we discuss an alternative to the Dirichlet distribution: the Beta-Liouville distribution. This distribution, which has only one additional parameter for the same vocabulary size, offers a more flexible correlation structure than the Dirichlet. Furthermore, we explore variational inference for the corresponding hierarchical Bayesian model. Leveraging the fact that the Beta-Liouville distribution, like the Dirichlet distribution, is a conjugate exponential family to the Multinomial, we are able to write the update equations for the \textsf{CAVI} algorithm and derive a stochastic version characterized by greater scalability and reduced computational load. 

Although empirical results are preliminary and further research is required, some meaningful insights have emerged. Notably, we compared the Dirichlet-Multinomial model with the Beta-Liouville Multinomial model using a non-informative hyperparameter setting. The Beta-Liouville prior was constructed to encompass the Dirichlet distribution as a special case. Our findings reveal a critical region in which the Beta-Liouville distribution outperforms the Dirichlet-Multinomial, specifically when word pairs exhibit a weaker negative correlation than that dictated by the Dirichlet prior. This effect is quantified by a hyperparameter, $\delta$, which measures divergence from the Dirichlet distribution, with the Beta-Liouville Multinomial model achieving optimal performance at $\delta < -0.20$.

A notable scenario arises when $\delta = 0$, corresponding to the Dirichlet distribution. Interestingly, results obtained via stochastic variational inference indicate that, in this case, the outcomes differ from those produced by the Dirichlet-Multinomial model. This discrepancy is not unexpected, as conventional strategies for maximizing the Evidence Lower Bound (ELBO) with respect to variational parameters within a Euclidean subspace inherently introduce an approximation, rather than capturing the true underlying geometry of the statistical problem. ELBO maximization is, in essence, a functional optimization problem defined on a statistical manifold parameterized by the variational distribution. Within this framework, information geometry identifies the Kullback-Leibler (KL) divergence as an appropriate Riemannian metric, accurately capturing the local geometric structure around a given member of the variational family \citep{amariNaturalGradientWorks1998}. Unlike the traditional \textsf{CAVI} algorithm, stochastic variational inference (\textsf{SVI}) effectively traverses the space of distributions by exploiting the correct gradient formulation and the direction of steepest descent induced by the KL-divergence \citep{hoffman_stochastic_2013}. The variational family defined for the Beta-Liouville Multinomial model differs from that of the Dirichlet-Multinomial model. This difference explains the apparent paradox: the \textsf{SVI} algorithm operates within two distinct spaces, leading to solutions that are not directly comparable, except in terms of the final partition quality.

The primary limitation to address in future work is the need for additional numerical experiments to better characterize the effects of different hyperparameter settings. Although we have provided preliminary insights here, these should not be regarded as conclusive. An intriguing direction for applications beyond text data analysis would be to explore mixtures of Beta-Liouville Multinomial regressions. In this framework, the hyperparameters of the Beta-Liouville distributions are not specified directly; instead, they are represented through a canonical link function as a function of a linear predictor, which in turn depends on a set of input variables. While this hierarchical structure offers significant potential, it also introduces immediate challenges. For instance, specifying a multivariate Gaussian prior for the parameters of the linear predictor results in a loss of conjugacy with the Beta-Liouville prior \citep{gormley_mixture_2018}. Thus, it remains uncertain whether and how the proposed variational algorithm can be effectively adapted in this context.

Another limitation of this work is that the examples presented were developed under the assumption that the number of components in the mixture was known. This, of course, is not the case in most real-world scenarios, and introduces additional complexity in the model selection phase. In principle, the \textsf{ELBO} provides a tight approximation to the marginal log-likelihood of the model for a fixed $G$, and could therefore be used to select the number of components when $G$ varies. However, in practice, we do not know how tight this lower bound is, and the variational gap between the \textsf{ELBO} and the marginal log-likelihood changes with $G$, complicating comparisons \citep{Murphy2013}. This makes it sensible to use information criteria, such as the Bayesian Information Criterion (BIC), appropriately modified for use within the MAP estimation framework provided by variational inference \citep{Fraley2002,Fraley2007}. It is evident that studying the interactions between the model selection procedures and the choice of the tuning parameter $\alpha$ is another crucial issue, particularly in terms of how this choice might lead to overfitting or underfitting with respect to the number of components in the mixture. Currently, this topic remains largely unexplored, and much work remains to be done.

Nevertheless, these preliminary results are promising. The use of the Beta-Liouville distribution as a prior for a Multinomial probability vector enriches the existing toolkits for computational analysis of short documents.

\backmatter

\begin{appendices}

\section{Expression of the ELBO}\label{sec:A1}
Terms \eqref{eq:ELBO1}-\eqref{eq:ELBO4} refer to $\operatorname{E}_{q_V}\left[\log  p(\pmb y_{1:n}, \pmb z_{1:n}, \pmb{\mathbbm P}, \pmb \lambda \vert \alpha_1, \ldots, \alpha_{p-1}, \alpha, \beta, \psi)\right]$. Terms \eqref{eq:ELBO5}-\eqref{eq:ELBO7} refer to $\operatorname{E}_{q_V}\left[\log q_V(\pmb z_{1:n}, \pmb{\mathbbm P}, \pmb \lambda\vert \pmb y_{1:n},  \pmb \gamma_{1:n}, \pmb \phi_{1:G}, \pmb \eta)\right]$:

\begin{align}
	\operatorname{ELBO}(q_V) & = \sum_{i=1}^n\sum_{g=1}^G\sum_{\ell=1}^{p-1} y_{i\ell}\gamma_{ig}\left\{\Psi(\phi_{g\ell})- \Psi(\sum_{\ell =1}^{p-1} \phi_{g\ell})+\Psi(\alpha)-\Psi(\alpha + \phi_{g\beta})\right\} \label{eq:ELBO1} \\
	& + \sum_{i=1}^n\sum_{g=1}^G y_{ip}\gamma_{ig} \left\{ \Psi(\phi_{g\beta}) - \Psi(\alpha + \phi_{g\beta})\right\} \nonumber \\
	& + \sum_{i=1}^n\sum_{g=1}^G \gamma_{ig}\left\{\Psi(\eta_g)- \Psi(\sum_{g=1}^G \eta_g)\right\} \label{eq:ELBO2}\\
	& + \sum_{g=1}^G \sum_{\ell=1}^{p-1} (\alpha_\ell - 1)\left\{\Psi(\phi_{g\ell})- \Psi(\sum_{\ell =1}^{p-1} \phi_{g\ell})+\Psi(\alpha)-\Psi(\alpha + \phi_{g\beta})\right\} \label{eq:ELBO3}\\
	& + \sum_{g=1}^G (\alpha - \alpha_0)\left\{ \Psi(\alpha) - \Psi(\alpha + \phi_{g\beta})\right\} \nonumber  \\
	& + \sum_{g=1}^G (\beta - 1)\left\{ \Psi(\phi_{g\beta}) - \Psi(\alpha + \phi_{g\beta})\right\} \nonumber \\
	& + \sum_{g=1}^G (\psi_g - 1) \left\{ \Psi(\eta_g) - \Psi(\sum_{g=1}^G \eta_g)\right\} \label{eq:ELBO4}\\
	& - \sum_{i=1}^n\sum_{g=1}^G \gamma_{ig}\log \gamma_{ig} \label{eq:ELBO5}\\
	& - \sum_{g=1}^G\log\Gamma(\phi_{g0}) + \sum_{g=1}^G\log B(\alpha, \phi_{g\beta}) + \sum_{g=1}^G\sum_{\ell=1}^{p-1} \log \Gamma(\phi_{g\ell}) \label{eq:ELBO6} \\
	& - \sum_{g=1}^G \sum_{\ell=1}^{p-1} (\phi_{g\ell} - 1)\left\{\Psi(\phi_{g\ell})- \Psi(\sum_{\ell =1}^{p-1} \phi_{g\ell})+\Psi(\alpha)-\Psi(\alpha + \phi_{g\beta})\right\} \nonumber \\
	& - \sum_{g=1}^G (\alpha - \phi_{g0})\left\{ \Psi(\alpha) - \Psi(\alpha + \phi_{g\beta})\right\} \nonumber  \\
	& - \sum_{g=1}^G (\phi_{g\beta} - 1)\left\{ \Psi(\phi_{g\beta}) - \Psi(\alpha + \phi_{g\beta})\right\} \nonumber \\
	& - \log\Gamma(\sum_{g=1}^G \eta_g) + \sum_{g=1}^G \log\Gamma(\eta_g)  \label{eq:ELBO7}\\
	& - \sum_{g=1}^G (\eta_g -1)\left\{\Psi(\eta_g) - \Psi(\sum_{g=1}^G \eta_g)\nonumber\right\},
\end{align}
where $\phi_{g0} = \sum_{\ell=1}^{p-1} \phi_{g\ell}$. The algebra involved in demonstrating the expression of ELBO is rather tedious and lengthy; therefore, we have omitted it due to space constraints. However, it is available upon request for any interested reader.




\end{appendices}


\bibliography{R1_apr25_ARXIV}

\end{document}